\PassOptionsToPackage{nopatch=footnote}{microtype}
\PassOptionsToPackage{hyperfootnotes=false}{hyperref}

\documentclass[sigconf,screen,nonacm]{acmart}
\renewcommand\footnotetextcopyrightpermission[1]{}
\usepackage{graphicx}
\usepackage{booktabs}
\usepackage{multirow}
\usepackage{array}
\usepackage{tabularx}
\usepackage{amsmath,amsfonts}
\usepackage{textcomp}
\usepackage{pifont}
\usepackage{algorithm}
\usepackage{algorithmicx}
\usepackage{algpseudocode}
\usepackage{balance}
\usepackage{longtable}
\usepackage{calc}
\usepackage{caption}
\usepackage{setspace}
\usepackage{placeins}
\usepackage{xcolor}
\usepackage{xspace}

\AtBeginDocument{%
  \hypersetup{%
    linkcolor=ACMDarkBlue,
    citecolor=ACMDarkBlue,
    urlcolor=ACMDarkBlue,
    filecolor=ACMDarkBlue}%
}

\AtBeginDocument{%
  \setlength{\abovedisplayskip}{5pt plus 1pt minus 1pt}%
  \setlength{\belowdisplayskip}{5pt plus 1pt minus 1pt}%
  \setlength{\abovedisplayshortskip}{4pt plus 1pt minus 1pt}%
  \setlength{\belowdisplayshortskip}{4pt plus 1pt minus 1pt}%
}
\makeatletter

\ExplSyntaxOn
\def\LT@output{%
  \ifnum\outputpenalty <-\@Mi
    \ifnum\outputpenalty > -\LT@end@pen
      \LT@err{floats~ and~ marginpars~ not~ allowed~ in~ a~ longtable}\@ehc
    \else
      \setbox\z@\vbox{\unvbox\@cclv}%
      \ifdim \ht\LT@lastfoot>\ht\LT@foot
        \dimen@\pagegoal
        \advance\dimen@\ht\LT@foot
        \advance\dimen@-\ht\LT@lastfoot
        \ifdim\dimen@<\ht\z@
          \setbox\@cclv\vbox{\unvbox\z@\copy\LT@foot\vfil}%
          \@makecol
          \@expl@@@mark@update@singlecol@structures@@
          \@outputpage
          \global\vsize\@colroom
          \setbox\z@\vbox{\box\LT@head}%
        \fi
      \fi
        \unvbox\z@\box\ifvoid\LT@lastfoot\LT@foot\else\LT@lastfoot\fi
        \UseTaggingSocket{tbl/longtable/foot}
    \fi
  \else
    \setbox\@cclv\vbox{\unvbox\@cclv\copy\LT@foot\vfil}%
    \UseTaggingSocket{tbl/longtable/foot}
    \@makecol
    \@expl@@@mark@update@singlecol@structures@@
    \@outputpage
      \global\vsize\@colroom
    \copy\LT@head\nobreak
  \fi}
\ExplSyntaxOff

\def\eg{\emph{e.g.}, } 
\def\ie{\emph{i.e.}, }

\newcommand{\ccmark}{\textcolor{green!80!black}{\ding{51}}}
\newcommand{\cxmark}{\textcolor{red}{\ding{55}}}

\newcommand{\figref}[1]{\hyperref[#1]{Figure~\ref*{#1}}}
\newcommand{\tabref}[1]{\hyperref[#1]{Table~\ref*{#1}}}
\renewcommand{\algref}[1]{\hyperref[#1]{Algorithm~\ref*{#1}}}
\newcommand{\eqnref}[1]{\hyperref[#1]{Equation~\ref*{#1}}}
\newcommand{\secref}[1]{\hyperref[#1]{Section~\ref*{#1}}}
\newcommand{\appref}[2]{\hyperlink{#1}{Appendix~#2}}
\newcommand{\apptabref}[2]{\hyperref[#1]{Table~#2}}
\newcommand{\AppendixAnchor}[1]{\phantomsection\hypertarget{#1}{}\label{#1}}
\makeatother
\newcommand{\AppendixSection}[2][]{%
  \par\addvspace{0.45\baselineskip}%
  \if\relax\detokenize{#1}\relax\else\AppendixAnchor{#1}\fi
  \noindent{\Large\bfseries #2}\par\nobreak\vskip 0.28\baselineskip}
\newcommand{\AppendixSubsection}[2][]{%
  \par\addvspace{0.35\baselineskip}%
  \if\relax\detokenize{#1}\relax\else\AppendixAnchor{#1}\fi
  \noindent{\large\bfseries #2}\par\nobreak\vskip 0.12\baselineskip}
\newcommand{\AppendixSubsubsection}[2][]{%
  \par\addvspace{0.22\baselineskip}%
  \if\relax\detokenize{#1}\relax\else\AppendixAnchor{#1}\fi
  \noindent{\normalsize\bfseries #2}\par\nobreak\vskip 0.02\baselineskip}
\newcommand{\AppendixTableCaption}[2][]{\par\vskip 0.45\baselineskip\if\relax\detokenize{#1}\relax\else\phantomsection\label{#1}\fi\begin{center}\begin{minipage}{0.94\linewidth}\centering\setstretch{1.08}\textbf{#2}\end{minipage}\end{center}\par\vskip -6pt\relax}

\long\def\AppendixCommandBlock#1{%
  \par\vskip 0.12\baselineskip
  \begingroup\leftskip=1.5em\noindent\emph{#1}\par\endgroup
  \vskip 0.12\baselineskip
}
\long\def\AppendixNumberedItem#1#2{%
  \par\noindent\hangindent=2.2em\hangafter=1\makebox[2.2em][r]{#1.\quad}#2\par
  \vskip 0.08\baselineskip
}
\newcolumntype{Y}{>{\centering\arraybackslash}X}
\newcolumntype{Z}{>{\raggedright\arraybackslash}X}
\newcolumntype{L}[1]{>{\raggedright\arraybackslash}p{#1}}
\newcolumntype{C}[1]{>{\centering\arraybackslash}p{#1}}
\definecolor{taskblue}{RGB}{52,152,219}
\definecolor{approvalgreen}{RGB}{39,174,96}
\definecolor{revisalred}{RGB}{231,76,60}
\definecolor{darkorange}{RGB}{211,84,0}
\definecolor{handoffpurple}{RGB}{142,68,173}
\algrenewcommand\algorithmicrequire{\textbf{Input:}}
\algrenewcommand\algorithmicensure{\textbf{Output:}}
\algnewcommand\algorithmicparameters{\textbf{Parameters:}}
\algnewcommand\Parameters{\item[\algorithmicparameters]}
\makeatletter
\global\@ACM@citypresenttrue
\global\@ACM@countrypresenttrue
\AtBeginDocument{%
  \fancypagestyle{standardpagestyle}{%
    \fancyhf{}%
    \fancyhead[C]{\@headfootfont\shorttitle}%
    \fancyfoot[C]{\if@ACM@printfolios\footnotesize\thepage\fi}%
  }%
  \pagestyle{standardpagestyle}%
}
\makeatother

\newcommand{\TableOne}{%
\begin{table*}[t]
\centering
\captionsetup{skip=6pt}
\caption{Comparison of LectūraAgents with existing multi-agent frameworks in this domain}
\label{tab:framework-comparison}
\renewcommand{\arraystretch}{1.22}
\setlength{\tabcolsep}{2.5pt}
\begin{tabularx}{\textwidth}{L{3.6cm}C{3.35cm}Y Y Y Y}
\toprule
\textbf{Framework} & \textbf{Teaching Modality} & \textbf{Embodied Agent(s)} & \textbf{Teaching Action Alignment} & \textbf{Personalization} & \textbf{Multi-Agent Collaboration}\\
\midrule
EduAgent~\cite{ref44} & Text & \cxmark & \cxmark & \cxmark & \cxmark\\
Agent4Edu~\cite{ref45} & Text (simulation) & \cxmark & \cxmark & \ccmark & \ccmark\\
EducationQ~\cite{ref46} & Text (simulation) & \cxmark & \cxmark & \cxmark & \cxmark\\
FACET~\cite{ref51} & Text & \cxmark & \cxmark & \ccmark & \cxmark\\
KELE~\cite{ref52} & Text & \cxmark & \cxmark & \cxmark & \cxmark\\
Instructional~Agents~\cite{ref49} & Text & \cxmark & \cxmark & \cxmark & \ccmark\\
EduPlanner~\cite{ref50} & Text & \cxmark & \cxmark & \ccmark & \ccmark\\
GenMentor~\cite{ref53} & Text & \cxmark & \cxmark & \ccmark & \ccmark\\
SimClass~\cite{ref47} & Text (simulation) & \cxmark & \cxmark & \ccmark & \ccmark\\
WikiHowAgent~\cite{ref48} & Text & \cxmark & \cxmark & \cxmark & \ccmark\\
\textbf{LectūraAgents} & \textbf{Multimodal (Embodied)} & \ccmark & \ccmark & \ccmark & \ccmark\\
\bottomrule
\end{tabularx}
\end{table*}}
\newcommand{\BoldSubsubsection}[1]{%
  \refstepcounter{subsubsection}%
  \par\addvspace{0.8\baselineskip}%
  \noindent{\bfseries\thesubsubsection\quad #1}\par\nobreak
  \vskip 0.25\baselineskip%
}

\newcommand{\FigureTwo}{%
\begin{figure*}[t]
\centering
\includegraphics[width=0.96\textwidth]{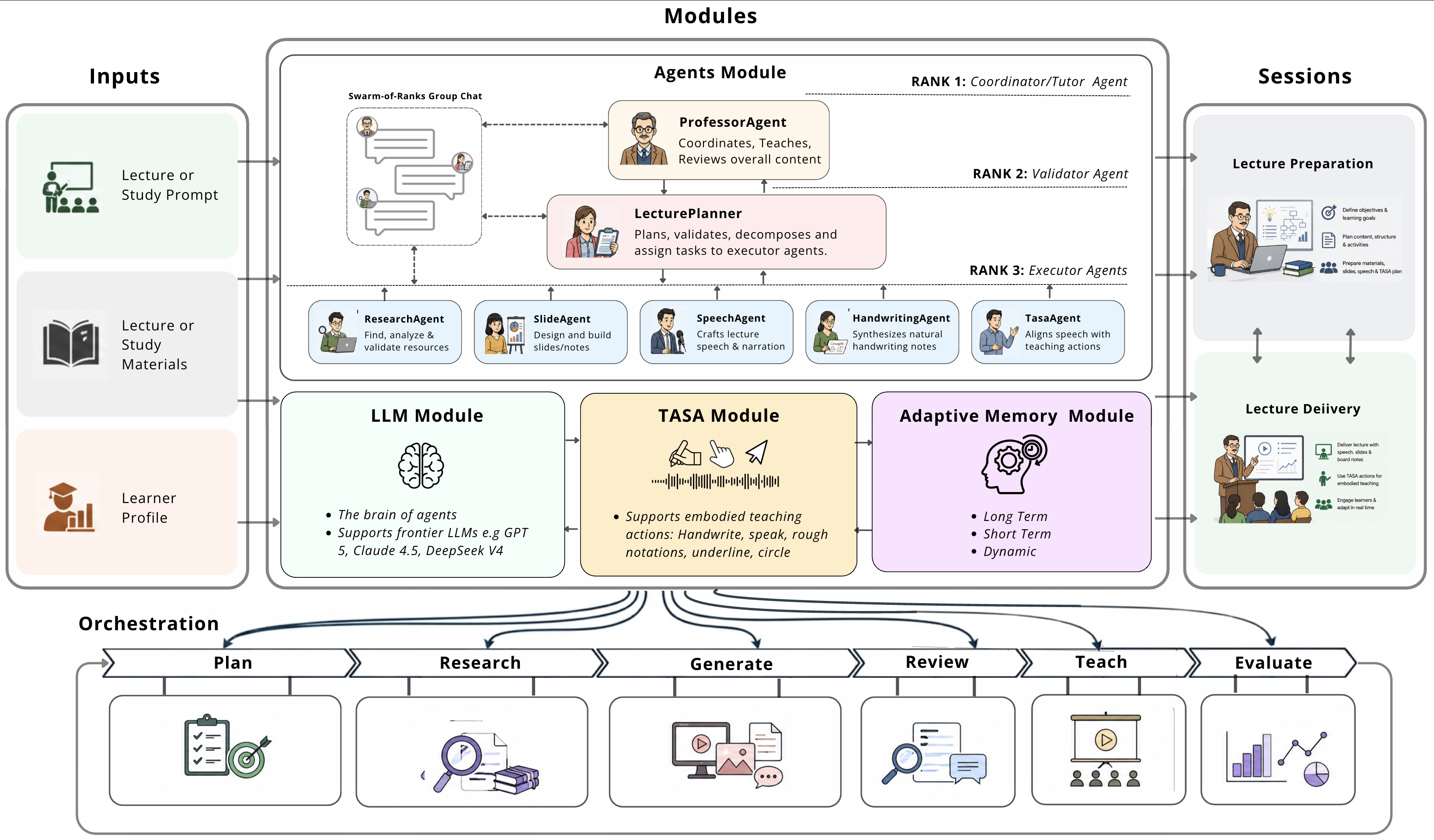}
\caption{LectūraAgents Architecture. The framework adopts a hierarchical multi-agent architecture, modelled after a professor--students' relationship. One in which a coordinator agent (or \emph{ProfessorAgent}) guides a collaborative team of validator and executor agents through planning, research, design, and delivery of personalized lecture contents. Multi-agent collaboration is mediated through an orchestration layer with group-chat communication that enables iterative planning, self-evaluation, and continuous refinement of generated materials. This architecture is supported by four interconnected modules: Agents, LLM, TASA, and Adaptive Memory.}
\Description{This figure is described in the caption.}
\label{fig:architecture}
\end{figure*}}

\newcommand{\FigureThree}{%
\begin{figure*}[t]
\centering
\includegraphics[width=\textwidth]{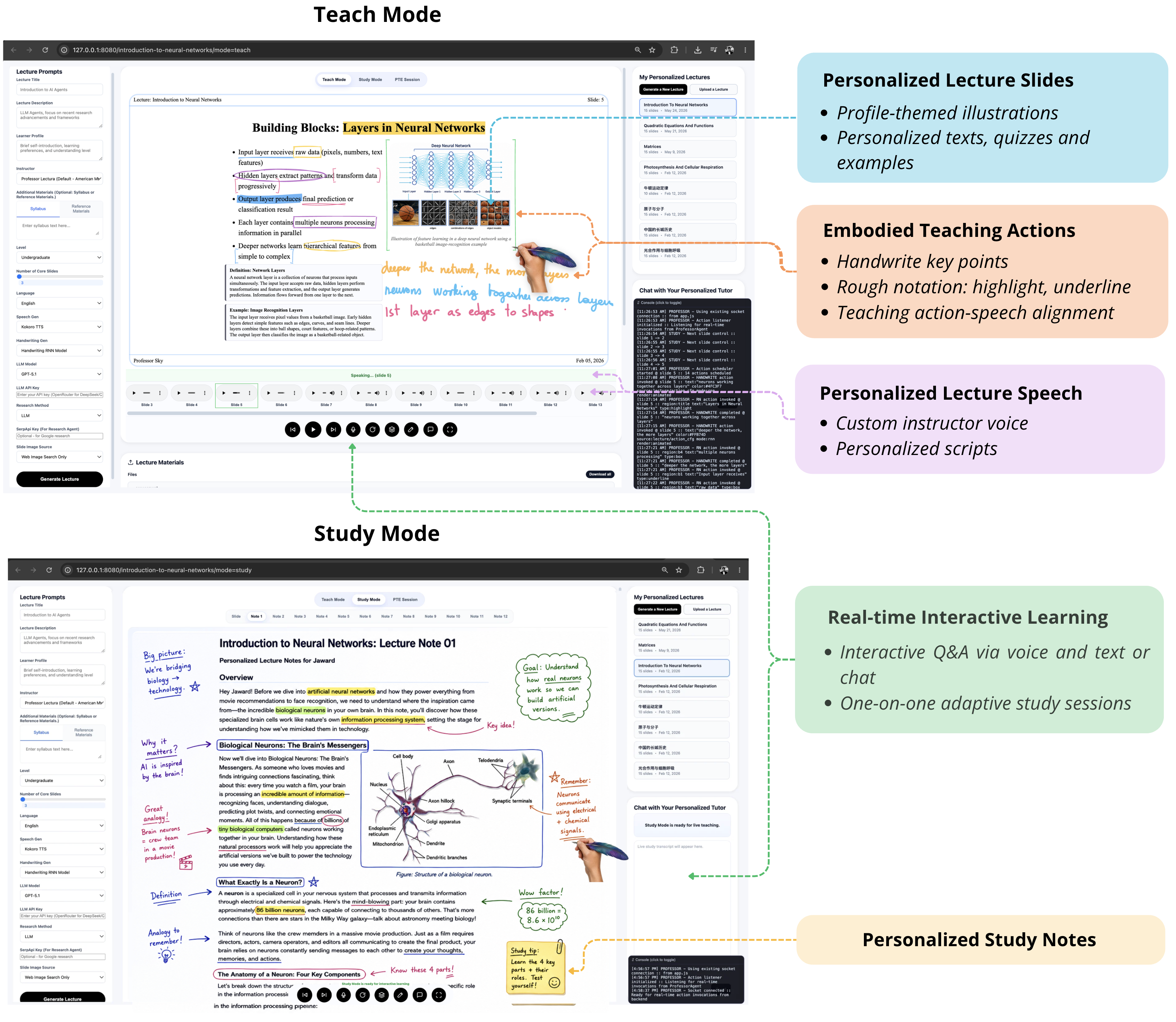}
\caption{An example of adaptive personalized learning experience with LectūraAgents}
\Description{This figure is described in the caption.}
\label{fig:adaptive-experience}
\end{figure*}}

\newcommand{\TableTwo}{%
\begin{table}[H]
\centering
\captionsetup{skip=6pt}
\caption{Message types for respective agents at different ranks}
\label{tab:message-types}
\renewcommand{\arraystretch}{1.0}
\setlength{\tabcolsep}{1pt}
\begin{tabularx}{\columnwidth}{C{1.15cm}C{2.9cm}Y}
\toprule
\textbf{Rank(s)} & \textbf{Message Types} & \textbf{Agents}\\
\midrule
1, 2 & \textcolor{taskblue}{[Task]}, \textcolor{approvalgreen}{[Approval]}, \textcolor{revisalred}{[Revisal]} & \emph{ProfessorAgent} and \emph{LecturePlanner}\\
1, 2, 3 & \textcolor{darkorange}{[TaskAcknowledged]}, \textcolor{taskblue}{[Progress]}, \textcolor{approvalgreen}{[TaskCompleted]} & \vspace*{.5em} All agents\\
3 & \textcolor{handoffpurple}{[Handoff]}, \textcolor{approvalgreen}{[RevisalSucceeded]}, \textcolor{revisalred}{[RevisalFailed]} & \emph{ResearchAgent, SlideAgent, ScriptAgent, SpeechAgent, TasaAgent}\\
\bottomrule
\end{tabularx}
\end{table}}

\newcommand{\AlgOne}{%
\begin{algorithm}[t]
\caption{Lecture Preparation Session}
\label{alg:prep}
\begin{tabularx}{\linewidth}{@{}lZ@{}}
\textbf{Input:} & Lecture Prompt $L_P$, Learner Profile $U$\\
\textbf{Parameters:} & Coordinator/validator agents $\{A_P,A_{LP}\}$; executor agents $E_A=\{A_R,A_S,A_{Sc},A_{Sp},A_T\}$;\\
& preparation plan $P=\{P_1,P_2,\ldots,P_n\}$; adaptive memory $M_A=\{M_s,M_L,M_d\}$;\\
& Swarm-of-Ranks group chat $G_{\text{chat}}$\\
\textbf{Output:} & Lecture artifacts $L_A=$ \{Plan, Slides, Script, Speech, TeachingActions, Notes, Assessments\}
\end{tabularx}
\begin{algorithmic}[1]
\State Initialize memory $\{M_s,M_L,M_d\}$ and agents $\{A_P,A_{LP}\}$
\State $A_P$ starts prep session and instantiates $G_{\text{chat}}$
\State $A_P$ debriefs $A_{LP}$ and requests lecture plan $P$
\Repeat
\State $A_{LP}$ drafts $P$ from $(L_P,U)\rightarrow M_d$
\State $A_P$ reviews $P$ and gives feedback $\rightarrow M_s$
\State $A_{LP}$ updates $P$ based on feedback $\rightarrow M_d$
\Until{$A_P$ approves $P$ or max iterations reached}
\State Initialize executor agents $E_A$
\For{each $P_i$ in $P$}
\State $A_{LP}$ debriefs $E_A$ on assigned tasks
\For{each executor $E_i$ in $E_A$}
\Repeat
\State $E_i$ plans and executes task
\State $E_i$ self-reflects and submits
\State $A_{LP}$ reviews task and gives feedback
\Until{$A_{LP}$ approves task or max iterations reached}
\EndFor
\EndFor
\State $A_{LP}$ submits artifacts $L_A$ for final review by $A_P$
\State $A_P$ reviews and validates $L_A$
\State \Return $L_A$
\end{algorithmic}
\end{algorithm}}

\newcommand{\FigureFour}{%
\begin{figure}[t]
\centering
\includegraphics[width=0.88\columnwidth]{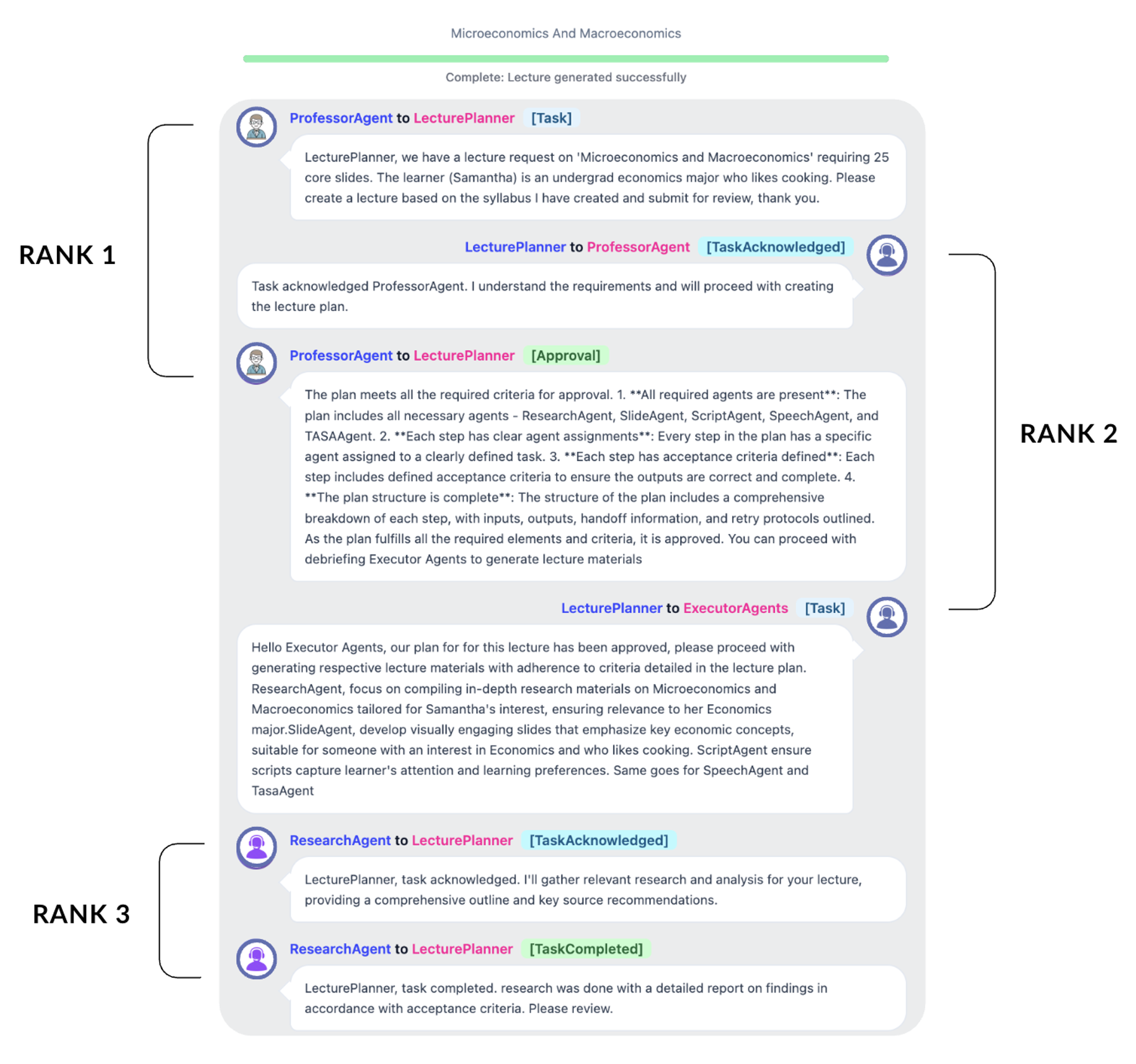}
\caption{Swarm-of-Ranks group chat}
\Description{This figure is described in the caption.}
\label{fig:swarm}
\end{figure}}

\newcommand{\FigureFive}{%
\begin{figure}[t]
\centering
\includegraphics[width=0.95\columnwidth]{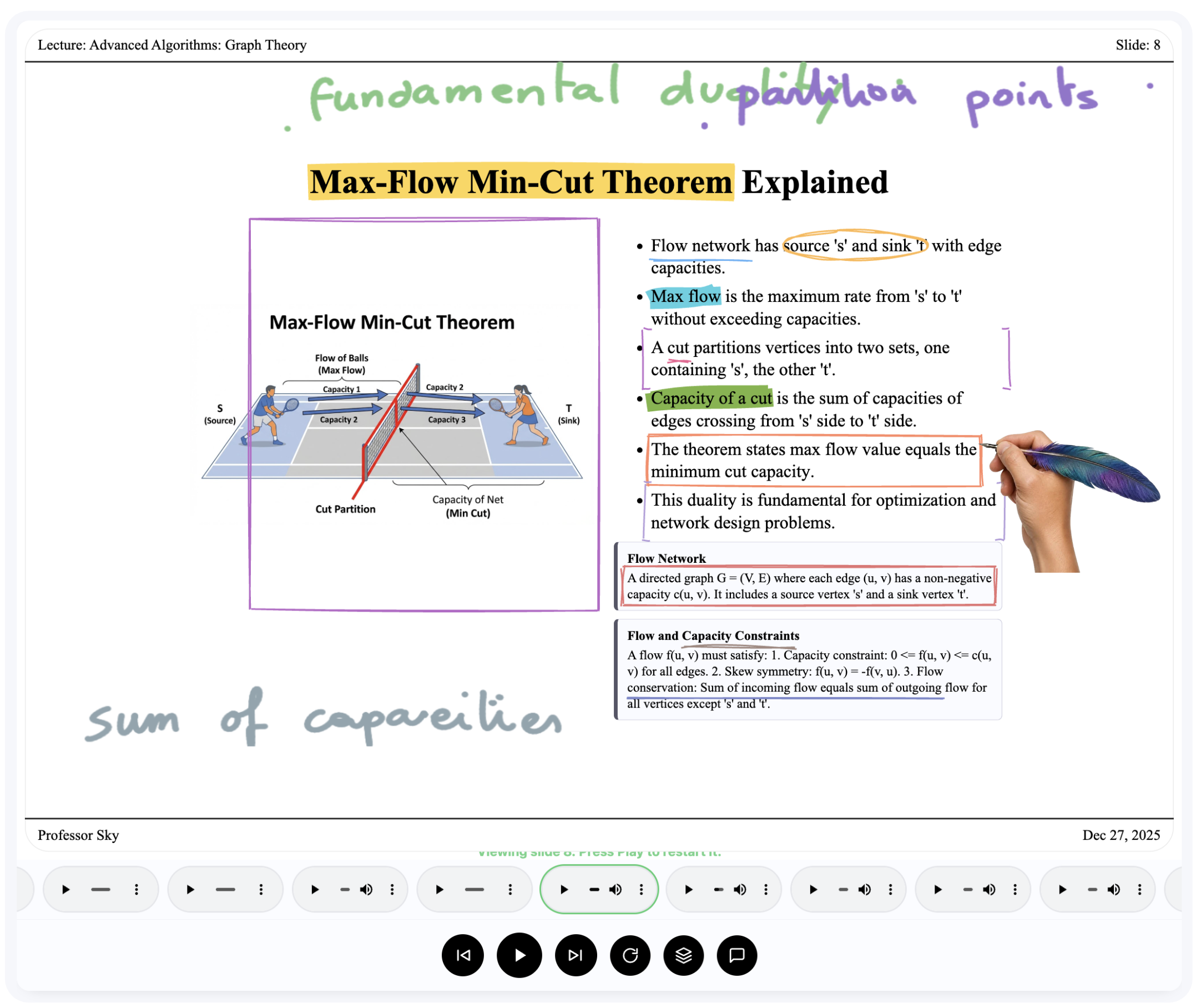}
\caption{Screenshot of a personalized slide for an undergraduate student whose favourite sport is tennis, with key concepts explained using tennis-themed visuals and embodied teaching actions over slide contents.}
\Description{This figure is described in the caption.}
\label{fig:personalized-slide}
\end{figure}}

\newcommand{\FigureSix}{%
\begin{figure}[H]
\centering
\includegraphics[width=0.94\columnwidth]{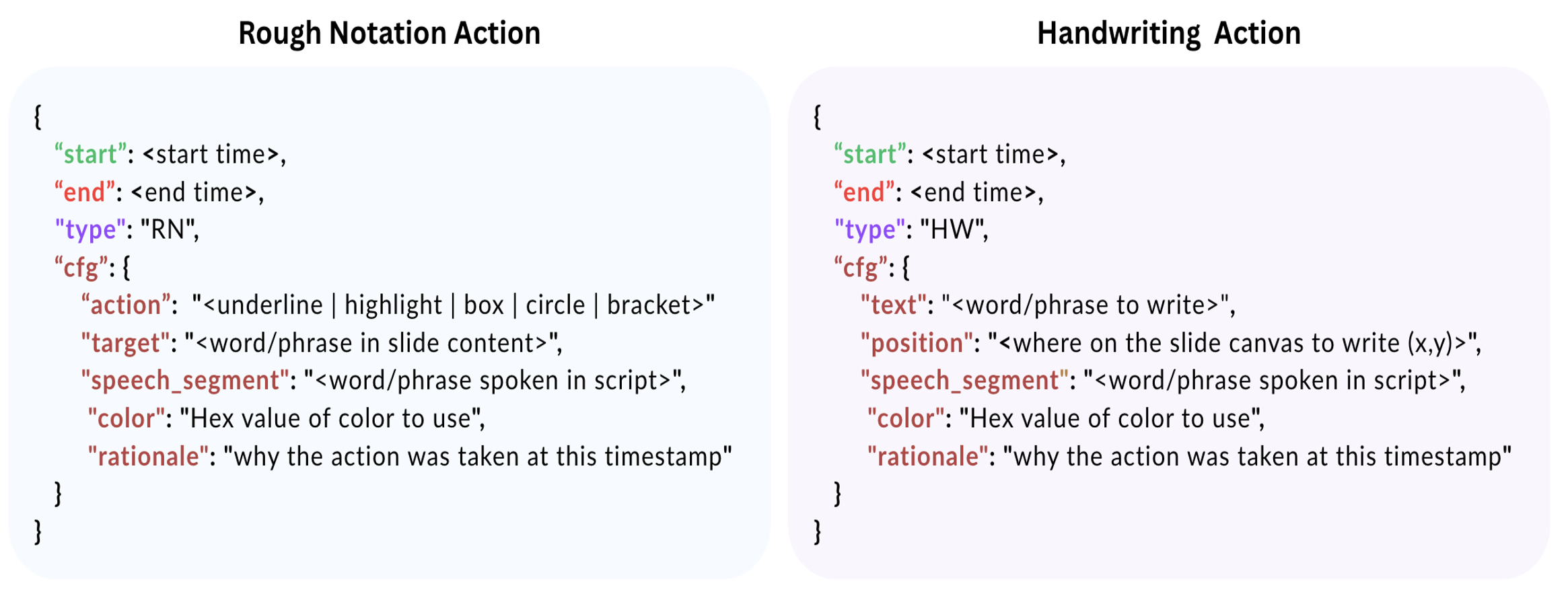}
\caption{Data structure for Rough Notation and Handwriting teaching actions in json.}
\Description{This figure is described in the caption.}
\label{fig:data-structure}
\end{figure}}

\newcommand{\FigureSeven}{%
\begin{figure*}[t]
\centering
\includegraphics[width=0.92\textwidth]{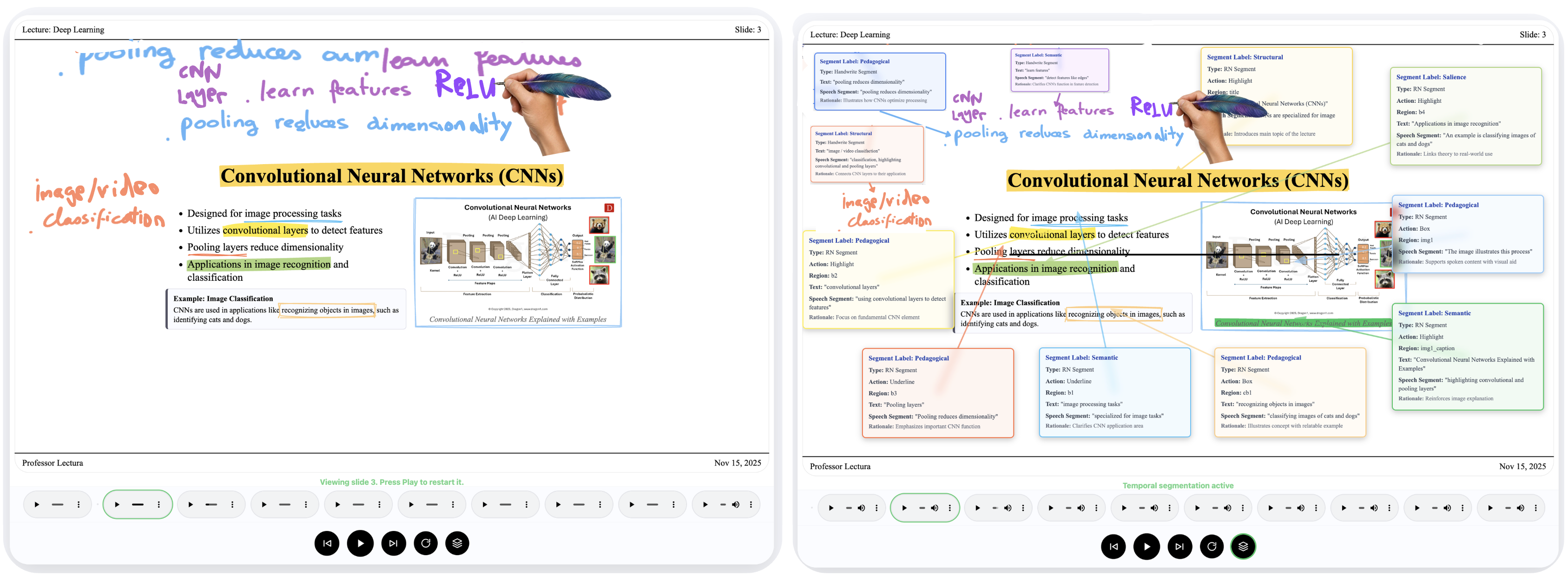}
\caption{The left slide shows already taken RN and HW teaching actions on a slide, while the right slide shows temporal semantic segmentation of slide contents with segment labels, action types, rationales, regions and their respective speech and script segments.}
\Description{This figure is described in the caption.}
\label{fig:semantic-segmentation}
\end{figure*}}

\newcommand{\AlgTwo}{%
\begin{algorithm}[t]
\caption{Teaching Action-Speech Alignment (TASA) Algorithm}
\label{alg:tasa}
\begin{tabularx}{\linewidth}{@{}lZ@{}}
\textbf{Input:} & Slides $S=\{S_1,S_2,\ldots,S_n\}$, scripts $S_c=\{S_{c1},S_{c2},\ldots,S_{cn}\}$, word-level speech timestamps $T_d$, learner profile $U$\\
\textbf{Parameters:} & \emph{TasaAgent}, regions $R$, labels \{Pedagogical, Personalized, Salient, Adaptive, Assessment\};\\
& RN and HW action types, salience data $\mathcal{H}$, dynamic memory $M_d$\\
\textbf{Output:} & $AS_{\text{seq}}$
\end{tabularx}
\begin{algorithmic}[1]
\State Initialize $S$, $S_c$, $T_d$, and $U$
\For{each slide $S_n$ in $S$}
\State Parse slide contents and identify regions $R$ in $S_n$
\State Analyze script for current slide $S_n$
\For{each region $R_n$ in $R$}
\State $L_n\gets$ assign segment label $L$ to region $R_n$
\State $S_{cn}\gets$ add appropriate speech segment
\State $segment_n\gets$ write segment data to $M_d$
\EndFor
\For{each $segment_n$ in slide $S_n$}
\State Analyze $segment_n$
\State $a_n\gets$ assign suitable action (RN or HW)
\State $r_n\gets$ give rationale for action
\State $\mathcal{H}\gets$ write salience heuristic data to $M_d$
\EndFor
\State $\mathcal{T}\gets$ save segmentation and analysis data to $M_L$
\EndFor
\State TasaAgent utilizes $\mathcal{T}$ to generate $AS_{\text{seq}}$
\State \Return $AS_{\text{seq}}$
\end{algorithmic}
\end{algorithm}}

\newcommand{\FigureEight}{%
\begin{figure}[H]
\centering
\includegraphics[width=1.0\columnwidth]{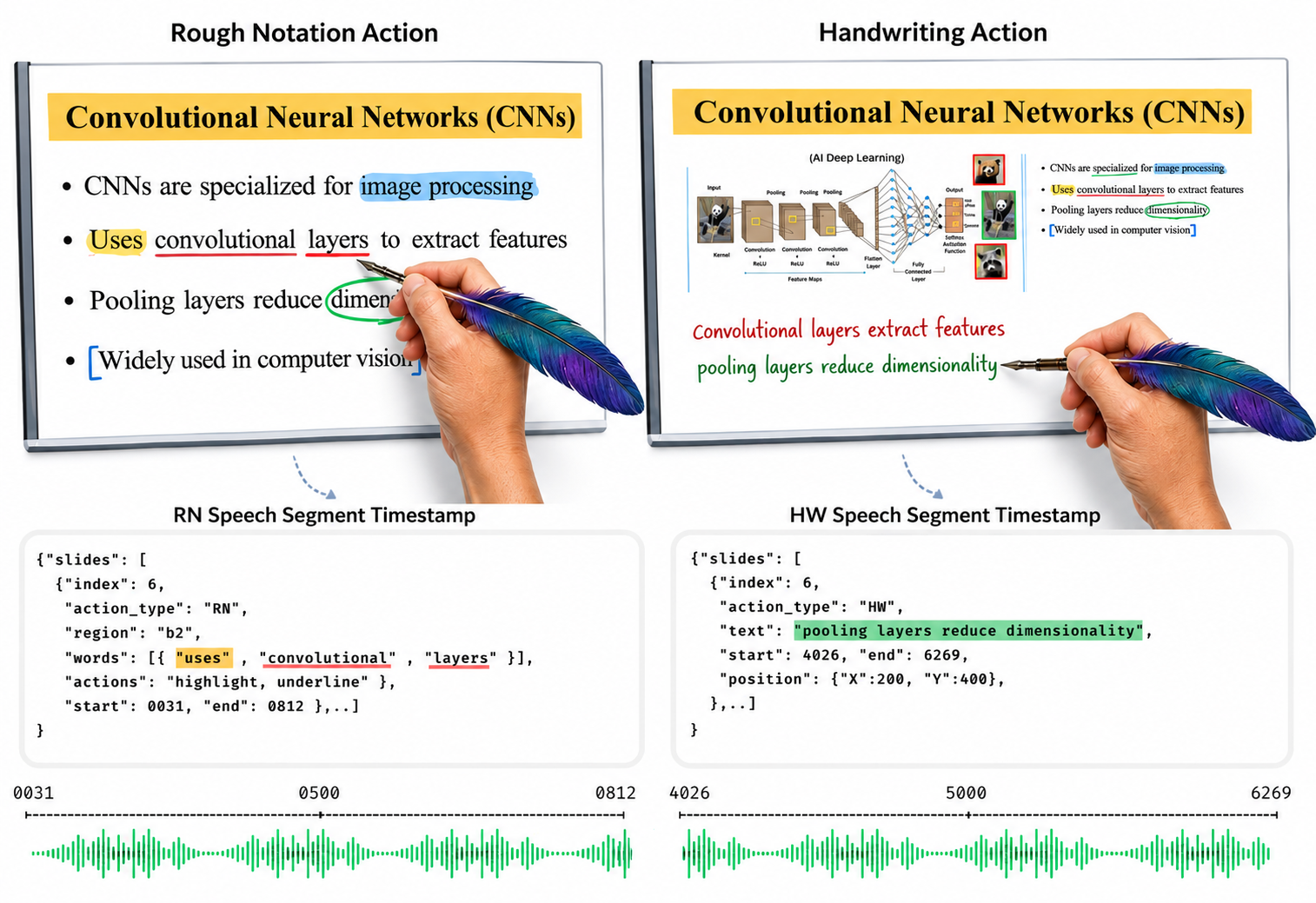}
\caption{Illustration of Embodied Teaching in LectūraAgents}
\Description{This figure is described in the caption.}
\label{fig:embodied-teaching}
\end{figure}}

\newcommand{\TableThree}{%
\begin{table}[t]
\centering
\captionsetup{skip=6pt}
\caption{Evaluation metrics and their respective rubrics}
\label{tab:evaluation-metrics}
\renewcommand{\arraystretch}{1.08}
\setlength{\tabcolsep}{2.5pt}
\begin{tabularx}{\columnwidth}{|L{2.85cm}|Z|}
\hline
\multicolumn{2}{|c|}{\textbf{Lecture Generation}}\\
\hline
\textbf{Evaluation Metric} & \textbf{Rubrics}\\
\hline
Lecture Content Quality (LCQ) & \emph{Accuracy, Clarity, Coherence, Cognitive Load, Syllabus Coverage, Instruction-following}\\
\hline
Personalization Quality (PQ) & \emph{Adaptive Emphasis, Preference Alignment, Engagement, Motivation, Tone/Style}\\
\hline
Assessment Quality (AQ) & \emph{Concept Coverage, Cognitive Appropriateness, Answer Validity; Rationale}\\
\hline
\multicolumn{2}{|c|}{\textbf{Lecture Delivery}}\\
\hline
Teaching Action Quality (TAQ) & \emph{Temporal Alignment, Accurate Handwriting Action, Accurate Rough Notation Action, Spatial Accuracy, Active Learning, Embodied Teaching}\\
\hline
\end{tabularx}
\end{table}}

\newcommand{\TableFour}{%
\begin{table*}[!t]
\centering
\captionsetup{skip=6pt}
\caption{(RQ 1) Evaluation of LectūraAgents across pedagogical metrics under frontier models}
\label{tab:frontier-models}
\renewcommand{\arraystretch}{1.12}
\setlength{\tabcolsep}{9pt}
\begin{tabular}{clccccc}
\toprule
\textbf{Rank} & \textbf{Model} & \textbf{LCQ (\%)} & \textbf{PQ (\%)} & \textbf{AQ (\%)} & \textbf{TAQ (\%)} & \textbf{AAR (\%)}\\
\midrule
1 & Gemini 3 Pro & 80.2 & 83.3 & 81.6 & 76.5 & \textbf{80.4}\\
2 & GPT-5.1 & 76.1 & 80.5 & 82.3 & 76.2 & \textbf{78.8}\\
3 & Claude 4.5 Sonnet & 72.4 & 78.6 & 76.2 & 80.4 & \textbf{76.9}\\
4 & Gemini 2.5 Pro & 70.5 & 75.2 & 80.1 & 72.3 & \textbf{74.5}\\
5 & DeepSeek V3.2 & 68.9 & 73.1 & 75.2 & 77.8 & \textbf{73.5}\\
6 & GPT-4o & 67.5 & 71.4 & 72.8 & 73.2 & \textbf{71.2}\\
7 & Qwen 3 Omni & 65.4 & 70.3 & 56.5 & 64.3 & \textbf{64.1}\\
\bottomrule
\end{tabular}
\end{table*}}

\newcommand{\FigureNine}{%
\begin{figure*}[!t]
\centering
\includegraphics[width=\textwidth]{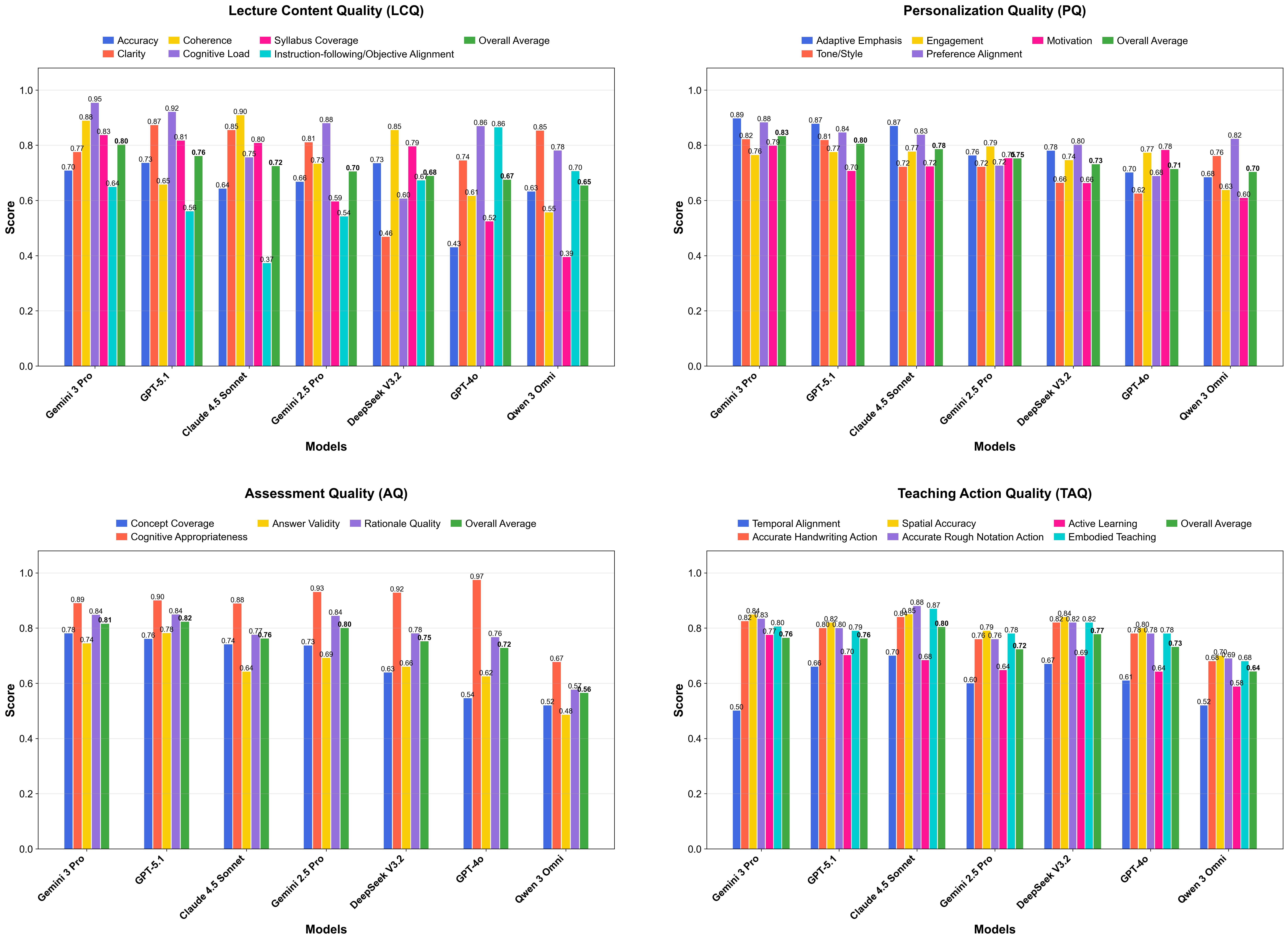}
\caption{(RQ1 and RQ2) Results across rubric dimensions for each evaluation metric under each frontier model.}
\Description{This figure is described in the caption.}
\label{fig:rubric-results}
\end{figure*}}

\newcommand{\FigureTen}{%
\begin{figure*}[t]
\centering
\begin{minipage}[t]{0.48\textwidth}
\centering
\includegraphics[width=\linewidth]{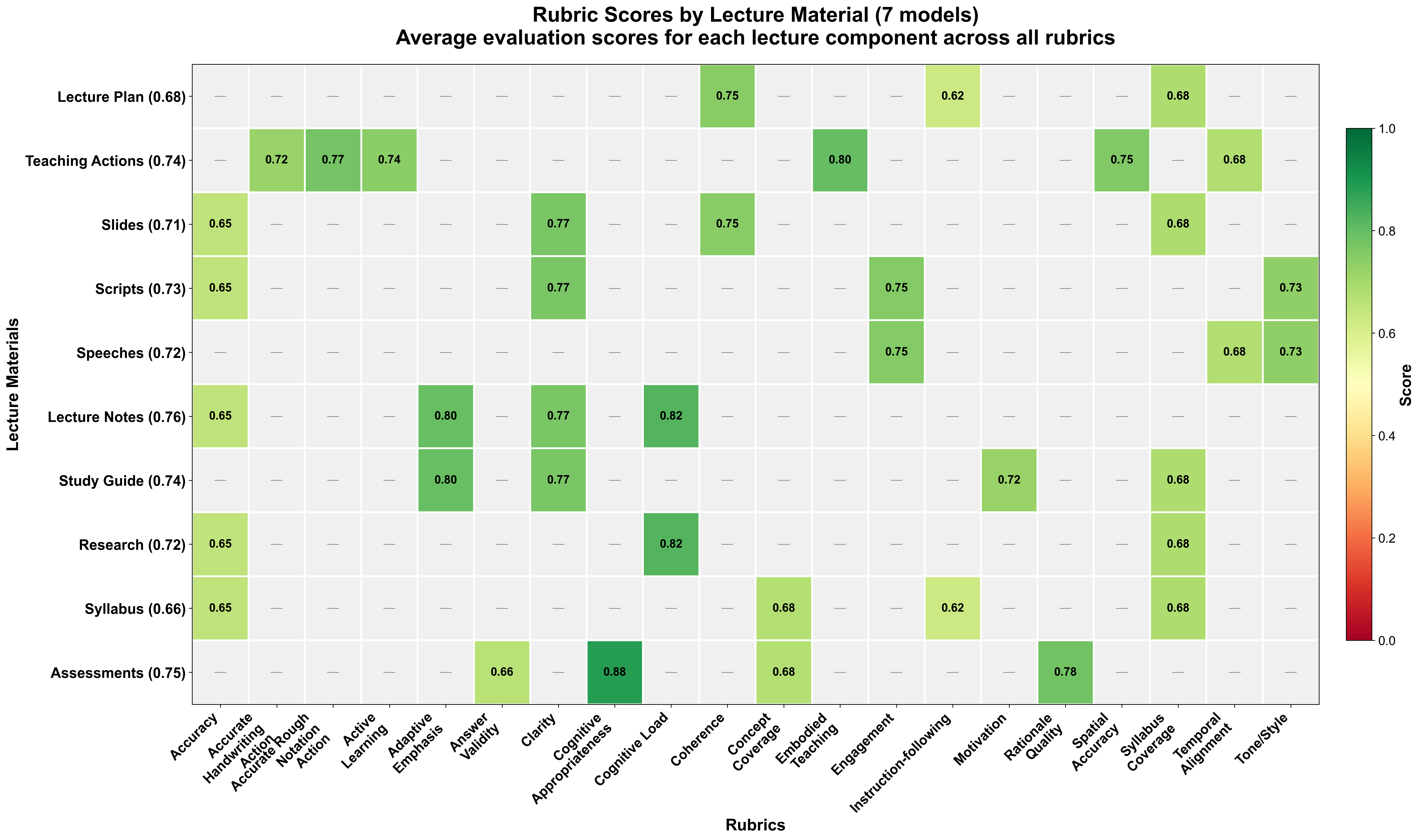}
\caption{(RQ2) Average distribution of Personalization Quality and Teaching Action Quality across diverse learning profiles at various academic levels.}
\Description{This figure is described in the caption.}
\label{fig:pq-taq-distribution}
\end{minipage}\hfill
\begin{minipage}[t]{0.48\textwidth}
\centering
\includegraphics[width=\linewidth]{fig11.png}
\caption{(RQ2) Overall average distribution of Lecture Content Quality scores across generated Lecture Materials from all models.}
\Description{This figure is described in the caption.}
\label{fig:lcq-distribution}
\end{minipage}
\end{figure*}}

\newcommand{\FigureEleven}{%
}

\newcommand{\TableFive}{%
\begin{table}[H]
\centering
\captionsetup{skip=6pt}
\caption{Performance comparison of LectūraAgents with existing related frameworks}
\label{tab:comparison}
\setlength{\tabcolsep}{2.5pt}
\begin{tabularx}{\columnwidth}{L{2.15cm}Y Y Y C{1.65cm}}
\toprule
& \multicolumn{4}{c}{\textbf{N (number of lectures) = 20}}\\
\cmidrule(lr){2-5}
\textbf{Framework / System} & \textbf{LCQ (\%)} & \textbf{PQ (\%)} & \textbf{AQ (\%)} & \textbf{Overall (\%)}\\
\midrule
Instructional Agents~\cite{ref49} & 52.1 & 53.2 & 51.4 & 52.2\\
GenMentor~\cite{ref53} & 50.8 & 64.6 & 46.6 & 54.0\\
Learn Your Way~\cite{ref3} & 58.9 & 60.1 & 62.5 & 60.5\\
\textbf{LectūraAgents} & \textbf{70.3} & \textbf{73.5} & \textbf{71.2} & \textbf{71.6}\\
\bottomrule
\end{tabularx}
\end{table}}

\newcommand{\TableSix}{%
\begin{table*}[t]
\centering
\captionsetup{skip=6pt}
\caption{Student responses to a survey given after assessment}
\label{tab:student-survey}
\renewcommand{\arraystretch}{1.35}
\setlength{\tabcolsep}{5pt}
\begin{tabularx}{\textwidth}{L{8.7cm}C{2.55cm}C{2.55cm}C{2.55cm}}
\toprule
\emph{To what extent do you agree or disagree with the following statements? \newline \% somewhat agree or strongly agree} & \textbf{LectūraAgents\newline N = 15} & \textbf{Learn Your Way\newline N = 15} & \textbf{Adobe Reader\newline N=15}\\
\midrule
\emph{I felt adequately prepared to complete the assessment after using today's educational tool.} & 95\% & 80\% & 72\%\\
\emph{I felt like today's educational tool helped me gain a good understanding of the topic.} & 100\% & 92\% & 65\%\\
\emph{I would like to use today's educational tool to support my learning needs in the future.} & 87\% & 73\% & 63\%\\
\emph{The educational tool I used today would make me more effective at learning compared to other educational tools I currently use at home or in school.} & 84\% & 67\% & 44\%\\
\bottomrule
\end{tabularx}
\end{table*}}

\newcommand{\FigureTwelve}{%
\begin{figure}[t]
\centering
\includegraphics[width=0.95\columnwidth]{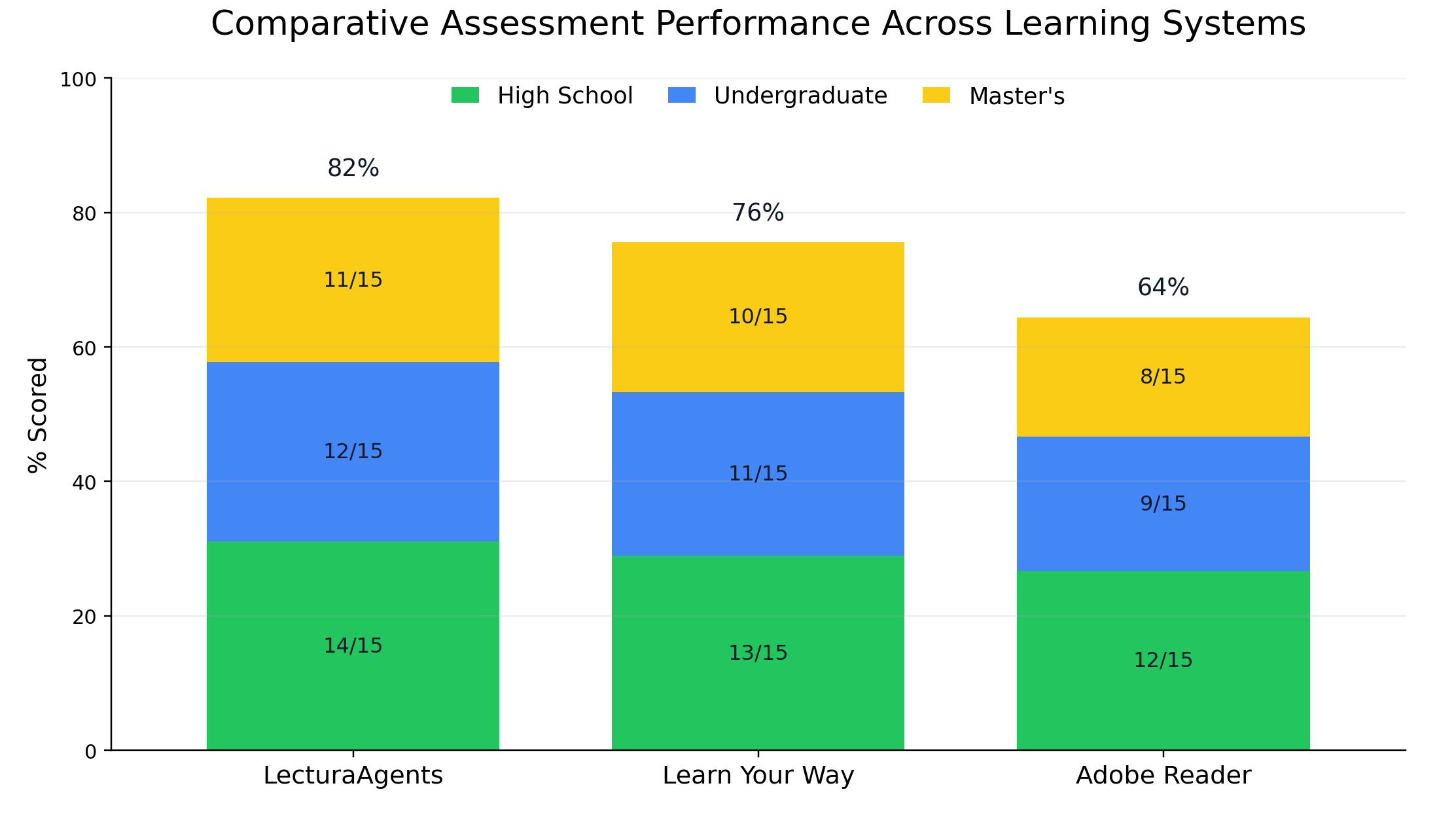}
\caption{Average scores from immediate assessment on topics learned using LectūraAgents, Learn Your Way, and Adobe reader.}
\Description{This figure is described in the caption.}
\label{fig:student-assessment}
\end{figure}
}

\title{LectūraAgents: A Multi-Agent Framework for Adaptive Personalized AI-Assisted Learning and Embodied Teaching}

\author{Jaward Sesay}
\affiliation{\institution{Beijing Institute of Technology}}
\author{Yue Yu\textsuperscript{\dag}}
\affiliation{\institution{Beijing Institute of Technology}}
\author{Siwei Dong\textsuperscript{*}}
\affiliation{\institution{Peking University}}
\author{Börje F. Karlsson\textsuperscript{*}}
\affiliation{\institution{Beijing Academy of Artificial Intelligence}}

\begin{teaserfigure}
\centering
\includegraphics[width=0.76\textwidth]{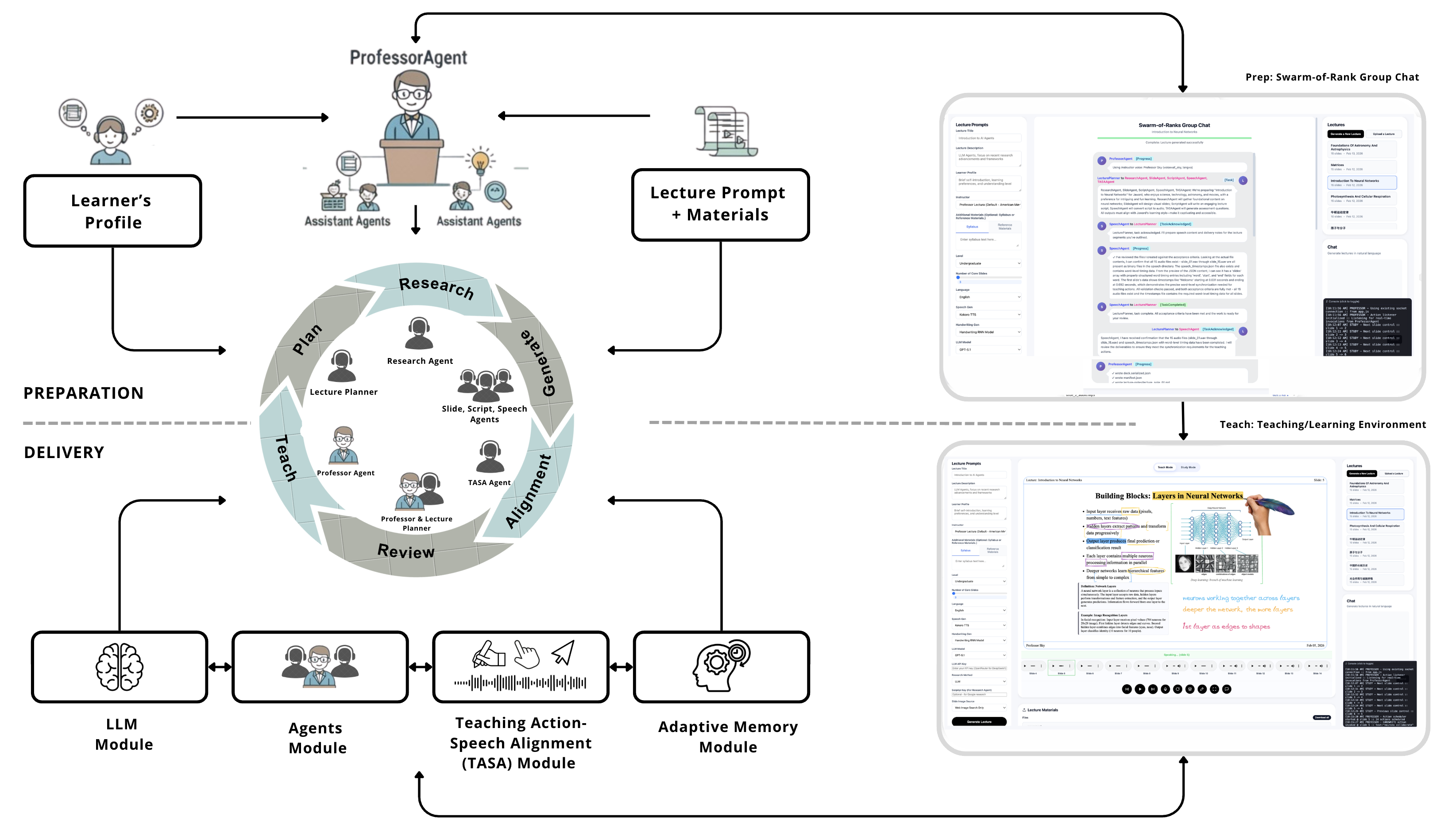}
\caption{Overview of LectūraAgents: a hierarchical multi-agent framework for end-to-end adaptive personalized learning experiences. Given a lecture prompt or learning materials and a learner profile, a \emph{ProfessorAgent} leads a collaborative team of specialized agents through research, planning, design, evaluation and embodied delivery of lecture and study contents that adapt to the individual learner. The framework provides students with access to real-time adaptive, personalized teaching and study sessions.}
\Description{This figure is described in the caption.}
\label{fig:overview}
\vspace{5pt}
\end{teaserfigure}

\begin{abstract}
Effective personalized AI-assisted learning demands learning systems that can not only generate accurate learner-specific educational materials, but also dynamically adapt their instruction to diverse learners. However, existing educational agent frameworks have primarily focused on lecture content automation and simulations, which often fall short of modelling multimodal and embodied instructional methods tailored for the individual learner. To this end, we propose LectūraAgents---a multi-agent framework that enables personalized learning through end-to-end adaptive embodied teaching. At its core, LectūraAgents mirrors a professor-student relationship, in which the \emph{ProfessorAgent} leads a collaborative team of specialized subordinate agents through research, planning, review, and embodied delivery of lecture contents that adapt to a learner's needs. The framework offers three main contributions: (1) a hierarchical multi-agent architecture for end-to-end personalized learning; (2) an adaptive embodied teaching mechanism, wherein the \emph{ProfessorAgent} executes visible and pedagogically motivated teaching actions (\eg handwrite, highlight, underline, etc.) over contents in a teaching environment while speaking; and (3) a Teaching Action-Speech Alignment (TASA) algorithm that employs salience-based heuristics and temporal semantic segmentation to generate coherent teaching action sequences aligned with learner profiles. We evaluate LectūraAgents on diverse courses at high school, undergraduate, and graduate levels using sample-specific rubric-based analysis; with generated lecture materials and teaching actions assessed and validated by expert educators. Experimental results show consistent gains in lecture content quality, embodied teaching quality, assessment, and personalization over existing approaches, positioning LectūraAgents as a pedagogically well-grounded framework for personalized learning at scale.
\end{abstract}

\thanks{\textsuperscript{\dag} Corresponding author: \href{mailto:yuyueanny@hotmail.com}{yuyueanny@hotmail.com}\\
\textsuperscript{*}\xspace\xspace Equal contributing authors\\
The dataset for this work is available at: \url{https://huggingface.co/datasets/Jaward/lectura-agents-data}}

\begin{document}
\maketitle

\section{Introduction}
\label{sec:introduction}

Adaptive personalized AI-assisted learning has emerged as a promising approach for tailoring instructions to individual learners, with studies reporting gains in motivation, engagement, and learning outcomes, especially in online educational settings~\cite{ref1,ref2,ref3}. However, contemporary personalized learning solutions and frameworks typically focus on adapting what is recommended, rather than how instructional content is delivered to the learner~\cite{ref4}. Research on embodied teaching shows that performing teaching actions (\eg writing, pointing or gesturing) during a lecture can help guide attention, foster conceptual understanding, and enhance overall learning outcomes~\cite{ref5,ref6,ref7}. These findings point to the need for personalized learning solutions that well integrate adaptive learning contents with embodied instructional delivery.

Recent frontier models demonstrate strong reasoning and agentic capabilities that have enabled planning, tool-use, and multi-step problem solving, opening new possibilities for applications in personalized learning~\cite{ref8,ref9,ref10,ref11}. This breakthrough has led to the exploration of LLM-powered agent frameworks for education, where specialized agents automate learning and teaching tasks to support students and educators~\cite{ref12}. Moreover, recent efforts have further demonstrated the potential of leveraging multiple agents to act as personal tutors and learning companions that provide on-demand teaching and learning support based on individual needs~\cite{ref13,ref14,ref15,ref16}.

However, the predominant focus of most related frameworks has been on simulations, where agents enact roles in virtual classrooms~\cite{ref17,ref18,ref19} simulate teacher--student dialogues to evaluate teaching behaviours and feedback strategies~\cite{ref20,ref21}, or coordinate agent workflows for generating personalized learning materials~\cite{ref22,ref23}. These are important proof of concepts, but their impact is limited to controlled virtual environments that do not capture the myriad nuances of adaptive personalized learning in real life scenarios. Other works have explored single-agent or prompt-engineered LLM tutoring systems~\cite{ref24,ref25,ref26} that generate explanations, feedback, or instructional contents, but without rigorous review or modelling of how such contents should be contextualized and adapted to diverse learning profiles. Few related works extend beyond these scopes to adopt a broader personalized instructional perspective that is often centred on automating course content generation~\cite{ref27,ref28}, which is primarily delivered in text-only modality, with no account for personalized embodied instructional delivery. Collectively, these systems offer valuable contributions to AI-assisted learning but remain fragmented in scope, lacking a unified model that connects personalized content generation with adaptive embodied delivery. Consequently, key pedagogical features, including coordinated lesson planning, iterative content review, embodied teaching, and alignment between teaching behaviour and learner needs, remain insufficiently addressed.

To address these limitations, we propose LectūraAgents, a hierarchical multi-agent framework for end-to-end personalized lecture generation and embodied lecture delivery. 

Our framework moves beyond simulations and static content generation, to managing the entire life cycle of a lecture (\ie from preparation to delivery, as shown in \figref{fig:overview}), while adapting to individual learning preferences. LectūraAgents offers three primary contributions:

\begin{enumerate}
\item[\textbf{1.}]
  \textbf{A hierarchical multi-agent architecture for end-to-end personalized learning:}~we propose the first multi-agent framework with end-to-end personalization for learning. It mirrors a professor--student relationship, where a ProfessorAgent coordinates specialized assistant agents (at different hierarchies) to plan, research, review, and create lecture contents tailored for the individual learner.
\item[\textbf{2.}]
  \textbf{A Teaching Action-Speech Alignment (TASA) algorithm:} a novel technique that uses LLM-based semantic analysis, temporal content segmentation, and salient heuristics to accurately align relevant teaching actions to regions or contents in a teaching environment (\eg over a slide).
\item[\textbf{3.}]
  \textbf{An embodied lecture delivery mechanism:} our framework enables a ProfessorAgent to perform visible, interpretable teaching actions (\eg highlight, handwrite, underline, etc.) directly over contents in the teaching environment (in our case, lecture slides) with a clear pedagogical rationale for each action taken.
\end{enumerate}

LectūraAgents decomposes personalized instruction into agents operating
at three hierarchies across two sessions: \textbf{Lecture Preparation}
and \textbf{Lecture Delivery}. In the preparation session, the
\emph{ProfessorAgent} leads a team of validator and executor agents
through planning, research, generation, and evaluation of lecture
artifacts. During the delivery or teaching session, the
\emph{ProfessorAgent} utilizes these artifacts to enact an embodied
teaching role, executing visible and pedagogically motivated teaching
actions on contents in the learning environment.

We conducted extensive evaluations of the framework on diverse courses
at high school, undergraduate, and graduate levels, assessing lecture
quality, teaching quality and personalization. Our experiments show that
LectūraAgents can produce high quality lecture artifacts, while
effectively adapting personalized teaching strategies to diverse learner
profiles through coherent embodied teaching action sequences.

\section{Related Work}
\label{sec:related-work}

\TableOne

\subsection{Adaptive Personalized AI-Assisted Learning}
\label{sec:adaptive-learning}

The idea of personalized learning predates LLMs and LLM agents.
Early theories of memory, such as Atkinson and Shiffrin's model
of how information is encoded and rehearsed \cite{ref29}
and Cowan's account of short-term and long-term memory capacities
\cite{ref30}, helped establish the cognitive foundations
for adapting instruction to the ways learners process and retain
information. These insights inspired models of personalized learning
that emphasized learner-centred pathways, individualized pacing and
tailored support. Before LLMs became widely adopted, deep learning
models were used in intelligent tutoring systems (ITS) to monitor
learners' performances, adjust task difficulty, and deliver
personalized feedback \cite{ref31,ref32,ref33}. Reviews show that AI-assisted personalized
learning has a positive impact on students' engagement and learning
outcome across diverse learning settings \cite{ref34,ref35,ref36}. More recent empirical
studies of AI-driven adaptive platforms in university and
language-learning contexts, report gains across performance,
satisfaction, and self-directed learning \cite{ref37,ref38,ref39}. Collectively, these
findings make clear the significance of adaptive personalized
learning, forming the foundations upon which our framework is built.

\subsection{LLM Agent Frameworks for Education}
\label{sec:llm-agent-frameworks}

Early works on LLM agents demonstrated how language models can plan, use tools, decompose tasks, and coordinate multi-step reasoning across multiple collaborating agents~\cite{ref40,ref41,ref42,ijcai2024autoagents}. These capabilities soon inspired educational multi-agent frameworks~\cite{ref43}. For instance, EduAgent~\cite{ref44} models diverse student personas using cognitive-science priors, Agent4Edu~\cite{ref45} simulates learner responses with memory-based generative agents, and EducationQ~\cite{ref46} stages multi-agent teacher-student-evaluator interactions to assess teaching behaviours. Similarly, systems like SimClass~\cite{ref47} and WikiHowAgent~\cite{ref48} extend simulation to classroom dynamics and procedural learning. Course-content automation then became a focus, with Instructional Agents~\cite{ref49} generating full course materials through role-based collaboration, and EduPlanner~\cite{ref50} iteratively refining lesson plans via evaluator--optimizer agent loops. More recent works have also introduced personalization: FACET~\cite{ref51} creates learner-adapted worksheets, KELE~\cite{ref52} provides concept-level enrichment and feedback, and GenMentor~\cite{ref53} builds personalized learning paths from learner goals. While these contributions demonstrate how multi-agent systems can enhance learning, they lack relevant capabilities (as summarized in \tabref{tab:framework-comparison}) that integrates personalized content generation with embodied instructional delivery.

\subsection{Embodied Teaching Agents}
\label{sec:embodied-teaching-agents}

Embodied teaching in digital settings refers to instructional methods that combine verbal instruction with spatial teaching actions (\eg writing, highlighting, underlining, or pointing) over learning contents in a virtual teaching environment. These actions help guide attention, reduce cognitive load, and support concept formation~\cite{ref54,ref55}. Earlier models like AutoTutor and its variations~\cite{ref56,ref57} demonstrated the benefit of animated pedagogical agents capable of conversational scaffolding. Recent systems have explored programmatic video-based approaches, for example, Xu et al.~\cite{ref58} explored how AI-generated lecture videos compare with real lectures, while AutoLectures~\cite{ref59} converts slides into narrated videos with highlight actions (using Levenshtein and LLM-based matching), and PASS~\cite{ref60} automated slide and speech generation from word documents. These efforts emphasize the importance of action-based instructional cues, but fall short of delivering a coherent end-to-end personalized, adaptive, and pedagogically informed embodied instruction.

\section{LectūraAgents}
\label{sec:lecturaagents}

\FigureTwo

We designed LectūraAgents to be both domain-specific and extensible given the nature of the problem we are trying to solve. The framework integrates planning, research, and pedagogical embodiment within a cohesive, end-to-end hierarchical architecture that supports both personalization and continual learning in tandem. As detailed in \figref{fig:architecture}, LectūraAgents consists of four interconnected modules:

\begin{itemize}
\item
  \textbf{LLM} -- The LLM module provides agents with access to frontier models (\eg GPT-5, Gemini 3 pro, Claude Sonnet 4, Deepseek v3.2, and Qwen 3) through their respective custom APIs. It serves as the brain behind agents, handling text, image, and speech modalities.
\item
  \textbf{Agent} -- This module holds the core logic for each agent's   role, capabilities, and tools. It also enables coordinated multi-agent collaboration through dynamically invoked actions for assigned tasks. The framework adopts a three-tier hierarchical collaborative mechanism with a lead coordinator agent managing a validator agent, who in turn manages executor agents for lecture content generation. To complete tasks, agents can execute a series of actions utilizing custom tools.
\item
  \textbf{TASA} -- The Teaching Action-Speech Alignment (TASA) module handles alignment between embodied teaching actions and their corresponding lecture speeches. It provides logic for salient heuristic analysis and temporal semantic segmentation, which help provide context when curating relevant teaching action sequences.
\item
  \textbf{Memory} -- This module implements short-term, long-term, and dynamic memories, which together allow agents to preserve context, track learner needs, and adapt their behaviour over time.
\end{itemize}
These four modules span across the framework's two main stages:
\emph{Lecture Preparation Session} and \emph{Lecture Delivery Session}. Moreover, as shown in \figref{fig:adaptive-experience}, the delivery session supports two modes:
\emph{Teach Mode}, which generates a new personalized lecture based on the learner's profile and provided learning materials, and \emph{Study Mode}, which allows learners to upload existing materials, such as notes, books or projects, and interact with the  \emph{ProfessorAgent} through real-time Q\&A.

\FigureThree

\subsection{Lecture Preparation Session}
\label{sec:lecture-preparation}

In this stage, the \emph{ProfessorAgent} leads a collaborative team of specialized agents through planning, research, alignment, review, and creation of personalized lecture artifacts (\eg lecture plan, slides, scripts, speech, teaching actions, notes, etc.). A quick overview of the entire process can be found in \algref{alg:prep}.

\textbf{Lecture Prompts and Configs.} Lecture preparation begins by processing the learner's prompts along with a range of configuration choices that define the scope, style, and preferences of the lecture. The prompt captures the lecture topic, its intended coverage, and the learner profile, so the framework can adapt content depth and learning preferences, while optional syllabus or reference materials help anchor the lecture to a course context or source material. Additional settings specify the instructor persona, target academic level (high school, undergraduate, masters, or PhD), language of instruction (which currently includes English, Chinese, French, or Spanish), and the approximate number of slides to be generated. Learners can also choose their preferred voice model, handwriting mode (either Handwriting RNN Model or Preset Font Handwriting), LLM model, and \emph{research} method (using Wikipedia or Google search). Together, these inputs provide the initial conditions that guide downstream multi-agent collaboration, planning, research, content generation, and embodied teaching. Our teaching and learning environment can be accessed via a browser (as shown in \figref{fig:adaptive-experience}) for easier entry of all inputs. Additional details on lecture prompts and configurations can be found in \appref{app:code-data}{B}.

\textbf{Multi-agent Collaboration.} When a lecture is prompted, the \emph{ProfessorAgent} first initiates the preparation session, creating a collaborative group chat named \emph{Swarm-of-Ranks Group Chat} (shown in \figref{fig:swarm}) -- where agents at different ranks collaborate to complete assigned tasks. In this group chat we have a coordinator (\emph{ProfessorAgent}), a validator (\emph{LecturePlanner}), and different executor (\emph{ResearchAgent, SlideAgent, ScriptAgent, SpeechAgent,} and \emph{Teaching Action-Speech Alignment agent} or \emph{TasaAgent}) agents. The coordinator agent (Rank 1) supervises the validator agent (Rank 2), who in turn manages executor agents at Rank 3. This hierarchical structure allows for efficient review and successful completion of assigned tasks. Agents communicate by sending messages in the group chat through a communication layer. There are nine message types: {[}Task{]}, {[}TaskAcknowledged{]}, {[}Progress{]}, {[}TaskCompleted{]}, {[}Approval{]}, {[}Revisal{]}, {[}Handoff{]}, {[}RevisalSucceeded{]}, and {[}RevisalFailed{]}. \tabref{tab:message-types} shows the message-types respective agents can send in the group chat.

\TableTwo

\textbf{Planning.} The lecture preparation process starts with planning, wherein the \emph{ProfessorAgent} instructs the \emph{LecturePlanner} to draft a lecture plan based on the requested lecture topic and learner profile. The \emph{LecturePlanner} first conducts preliminary research on the topic, then writes a detailed plan, and submits it for review and approval by the \emph{ProfessorAgent}. The plan contains lecture metadata, learner profile, and detailed descriptions of tasks for each executor agent with respective criteria for completing assigned tasks. Once the plan is approved, the \emph{LecturePlanner} then instructs and coordinates executor agents to generate lecture contents based on the plan. Subsequent preparation stages will involve sequential execution of tasks by executor agents and iterative validation by the \emph{LecturePlanner}.

\textbf{Generation.} This stage starts with the \emph{SlideAgent}, which is tasked with designing each slide (in HTML format), using a custom slide builder tool, and generating respective contents based on structural and pedagogical criteria from the lecture plan. Each slide is designed to support contents in text, image, video, and speech modalities, via structured content blocks. Slide images can be either generated or sourced online via web search. Next, the \emph{ScriptAgent}
utilizes the generated slide contents (along with lecture plan and research findings) to create a personalized and pedagogically informed script for each created slide. Scripts are conditioned to capture the
learner's attention, level of understanding, and learning preferences. Finally, the scripts are then passed on to the \emph{SpeechAgent} which performs speech synthesis, converts scripts to speech (in the learner's desired instructor voice), and creates word-level timestamps for each speech action using Whisper ASR~\cite{ref66}. These artifacts will later be used during alignment and review.

\textbf{Alignment.} Given the generated speech timestamps, scripts, slide contents, and learner profile, the  \emph{TasaAgent} first performs a preliminary teaching action analysis using segmentation and salient heuristic tools in the TASA module. This analysis starts with the temporal semantic segmentation of slide contents and scripts to identify segments that should receive teaching actions; it then applies salience-based heuristics to provide rationale for each teaching action application.

Currently, LectūraAgents supports two kinds of teaching actions: Rough Notation (RN), \eg highlight, underline, circle, box, etc., and Handwriting actions (HW), \ie writing down key points in natural human-like handwriting style, while speaking. This analysis results are then added to the agent's context when mapping pedagogical teaching actions to contents in the slide teaching environment. The \emph{ProfessorAgent} will later utilize the resulting teaching action sequences during embodied teaching in the lecture delivery session.

\AlgOne
\FigureFour

\textbf{Self-reflection.} In addition to the hierarchical review mechanism present in multi-agent collaboration, we ensure each agent self-reflects on any completed tasks to find and fix issues before submitting results for review by the validator agent. They do this by first reviewing completed tasks, then self-validating them against required criteria detailed in the lecture plan.

\textbf{Personalization.} We ensure personalization across all generated lecture contents---slides, images, quizzes, lecture notes, scripts and teaching actions---by conditioning generation on the learner's profile, learning preferences, and usage history in memory. For example, slide contents, as shown in \figref{fig:personalized-slide}, are adapted to the learner's interests by framing concepts around a favourite sport or hobby, or can be tailored into an easier-to-follow learning path (\eg more scaffolding or simpler analogies) when the student profile indicates lower prior knowledge. Slide images are generated to match the same themes and difficulty level, while quizzes are personalized in both content and phrasing to assess understanding using familiar scenarios. The resulting notes, scripts, and teaching actions mirror these choices to ensure a coherent, learner-specific narrative throughout the lecture.

\FigureFive

\textbf{Review.} Finally in this session, generated lecture artifacts are assembled by the \emph{LecturePlanner} and submitted to the \emph{ProfessorAgent} for final review. During review the \emph{ProfessorAgent} again validates lecture artifacts based on lecture content quality, teaching quality, action alignment, and personalization. Once review is successful, the  \emph{ProfessorAgent} agent then takes on the role of teacher in the subsequent lecture delivery session.

\subsection{Lecture Delivery Session}
\label{sec:lecture-delivery}

During this stage, the \emph{ProfessorAgent} assumes the role of an embodied instructor that executes pedagogical teaching actions in the slide environment using lecture artifacts from the lecture preparation session. In this work, we define a \emph{\textbf{Teaching Action}} as a semantically bounded, visually interpretable and pedagogically motivated operation performed by the \emph{ProfessorAgent} over contents in the teaching environment, while speaking. Each action comes with a
rationale for why it was taken at a particular time. We experiment with two types of teaching actions:

\begin{enumerate}
\def\labelenumi{\arabic{enumi}.}
\item[\textbf{1.}]
  \textbf{Rough Notations (RN):} These are actions that involve marking or emphasizing existing contents on the slide. Examples include highlighting key terms, underlining important phrases, circling diagrams, or boxing critical points. RN actions are used to draw the learner's attention to specific areas of the slide that are relevant to the current topic being discussed. For improved user experience, we make use of a hand-drawn annotation library~\cite{ref61} that simulates human-like rough notations for these actions.

\item[\textbf{2.}]
  \textbf{Handwriting (HW):} These actions involve writing new information directly onto the slide canvas in a natural, human-like handwriting style. This can include jotting down definitions, drawing diagrams, or annotating existing content. HW actions serve to reinforce learning by actively engaging the learner with newly introduced concepts during the lecture. We utilize both a handwriting recurrent neural network model based on Graves~\cite{ref62} and a preset font-based handwriting synthesis for this teaching action.
\end{enumerate}
These actions undergo preliminary review, analysis, and alignment using our proposed Teaching Action-Speech Alignment (TASA) algorithm,
summarized in \algref{alg:tasa}.

\textbf{Teaching Action--Speech Alignment (TASA) Algorithm}. TASA uses a combination of LLM-based salience heuristics analysis and temporal semantic segmentation to help guide the \emph{TasaAgent} with prospective relevant teaching action sequences. The agent's objective is to emit an ordered list of pedagogically informed teaching action-speech sequences \(AS_{\mathrm{seq}}=\{S_1[a_1,a_2,\ldots,a_n],\ldots,S_n[a_1,a_2,\ldots,a_n]\}\), for each slide \(S_{n}\), where each action \(a_{n}\) is given by:
\begin{equation}
\label{eq:action-structure}
a_n = \{\mathrm{actiontype}_n,\mathrm{start}_n,\mathrm{end}_n,\mathrm{cfg}_n\}
\end{equation}
\(action{type}_{n}\) can be either RN or HW, (\({start}_{n},{end}_{n}\)) gives the duration for the action, and \({cfg}_{n}\) holds additional metadata or configuration specific to the action type, as illustrated in \figref{fig:data-structure}.

\FigureSix

\FigureSeven

\textbf{Temporal Semantic Segmentation.} Before performing salience heuristics analysis, we first segment slide contents and speech semantically (see \figref{fig:semantic-segmentation}), in order to augment our agent's context for better teaching action sequences. Segment labels include \emph{Pedagogical, Personalized, Salient, Adaptive}, and \emph{Assessment}, each of which helps provide insight into the kind of teaching actions to apply. For each slide region \({region}_{n}\) \ensuremath{\in} \(R_{s}\) and corresponding speech segment with label \({label}_{n}\), the TasaAgent creates a segment \({segment}_{n}\) given by:
\begin{equation}
\label{eq:segment}
\mathrm{segment}_n = \{\mathrm{label}_n,\mathrm{region}_n,\mathrm{speech\_segment}_n\}
\end{equation}
specifically, for each candidate segment \({segment}_{n}\) in a slide \(S_{n}\), TASA analyses the segment data and assigns a suitable teaching action along with a rationale \(r_{n}\) in natural language, explaining why this action is appropriate for that specific region. The final heuristics analysis data for a given slide is recorded as:
\begin{equation}
\label{eq:heuristics}
\mathcal{H}(S_n)=\{\mathrm{segment}_n,a_n,r_n\}
\end{equation}
which provides the \emph{TasaAgent} with a structured context when generating the resulting teaching action-speech sequences \emph{AS\textsubscript{seq}}.

\textbf{Embodied Teaching.} Given the generated teaching action-speech sequences, the \emph{ProfessorAgent} dynamically schedules and invokes respective teaching action functions over regions in the slide environment (in sequence), while speaking. Each action function is tied to a specific speech segment (with word-level timestamps) and applies a targeted visual operation such as handwriting, highlighting, circling, or underlining, directly on the corresponding slide region, as illustrated in \figref{fig:embodied-teaching}.

To ensure accurate and realistic embodiment, the agent is provided with a discrete world view of the slide environment and its contents, while using a 3D quill-holding hand to execute the embodied teaching actions with precise spatial targeting of regions and their corresponding action types. As a result, embodied teaching actions like handwriting, highlighting, circling, etc., are executed in a natural, interpretable, and pedagogically grounded manner that closely mirrors human instructional behaviour.

\FigureEight
\AlgTwo

\section{Experiments}
\label{sec:experiments}

We conducted extensive quantitative and qualitative evaluations of LectūraAgents through diverse experiments, assessing the framework's performance on the following pedagogical metrics: lecture content quality, teaching quality, assessment, and personalization. Our main goal is to provide answers to two fundamental research questions:

\begin{enumerate}
\def\labelenumi{\arabic{enumi}.}
\item[\textbf{1.}]
  \textbf{RQ1}: How does leveraging an adaptive hierarchical multi-agent architecture create high-quality personalized lecture contents that align with diverse learning profiles?

\item[\textbf{2.}]
  \textbf{RQ2}: How can an embodied tutor agent utilize generated materials to execute coherent, visual, and pedagogically informed teaching actions in a teaching environment (\eg lecture slides
  presentation)?
\end{enumerate}

\subsection{Experiment Setup}
\label{sec:experiment-setup}

The experiments were designed to assess the framework from end-to-end,
evaluating both personalized lecture generation and embodied teaching
capabilities. We start by performing pedagogical evaluation on 280
personalized lectures generated using the framework under the seven
frontier models reported in \tabref{tab:frontier-models}. For each model, we generate 40
lectures, with 10 lectures for each academic level, using the same
prompts, learner profiles, and text-to-speech model (Kokoro TTS~\cite{ref65}) to
ensure a fair comparison. Details on these lectures can be found in
\appref{app:pedagogical-eval}{A.2.2}. We worked with five expert educators, including subject
teachers and university instructors with experience in curriculum design
and instructional assessment, to define pedagogical rubrics grounded in
recognized instructional quality standards \cite{ref63},
as summarized in \tabref{tab:evaluation-metrics}. Additional details on the recruitment of these
experts can be found in \appref{app:expert-recruitment}{A.2.4}. We then adopted the evaluation
method in TutorBench~\cite{ref64}, with
scoring primarily done by the expert educators in order to avoid induced
bias from an LLM judge. Thus, for a j-th lecture, the framework's
overall performance score for each session under a given model or
baseline framework, is computed as the weighted average of all passed
rubric criteria \({AAR}_{w}^{j}\), given by:
\begin{equation}
\label{eq:aar}
{AAR}_{w}^{j} = \frac{\sum_{i = 1}^{N_j} w_i^j \cdot \mathbf{1}_{r_i^j}}{\sum_{i = 1}^{N_j} w_i^j \cdot \mathbf{1}_{w_i^j > 0}}
\end{equation}
where \(N_{j}\) is the number of rubric criteria for the j-th lecture,
\(w_{i}^{j}\)\ensuremath{\in} \{-5, -3, -1, 0, +1, +3, +5\}, is the weight assigned to
the i-th criterion,

and \(r_{i}^{j}\) \ensuremath{\in} \{0,1\} indicates whether criterion \emph{i} is
satisfied. When a criterion is satisfied \(r_{i}^{j} = 1\), it
contributes a positive reward of +5, +3, or +1, corresponding to it
being a highly desirable, desirable and important, or nice-to-have
behaviour, respectively. When a criterion is not satisfied
\(r_{i}^{j} = 0\), it is explicitly treated as a failure state and
contributes a non-positive score, spanning a 0, -1, -3, and -5 range: 0
denotes the lowest-severity failure (no credit), -1 is used for a minor
failure, -3 for a moderate failure, and -5 represents a critical failure
(highly undesirable behaviour).

\TableThree

\TableFour
\FigureNine

\BoldSubsubsection{Lecture Generation Evaluation}

\noindent Here, we evaluate LectūraAgents as a personalized lecture content generator. For each
model, we generated 40 personalized lectures covering maths, science,
engineering, art, and history, with 10 lectures each for high school,
undergraduate, master's, and PhD learning profiles. Topics were randomly
selected with emphasis on science subjects. Each lecture targeted one
individual learner profile, covering learners aged 13--35, with profiles
varying by academic level, prior knowledge, learning goals, learning
style, and expected difficulty. The resulting output after generation
contains the following lecture artifacts: a detailed lecture plan, a
research report, syllabus, learner profile, 15 slides with images,
per-slide scripts, lecture speeches, personalized lecture notes and study
guide, teaching actions, teaching action--speech alignment, and
assessments.

\emph{Evaluation Metrics.} Using expert-defined rubrics detailed in
\tabref{tab:evaluation-metrics}, we assess the framework's personalized lecture content
generation capability across three main evaluation metrics: Lecture
Content Quality (LCQ), Personalization Quality (PQ), and Assessment
Quality (AQ). LCQ measures accuracy, clarity, coherence, cognitive load,
and instruction-following rubric dimensions. PQ evaluates adaptation to
a learner profile (adaptive emphasis) and learning preferences
(preference alignment), engagement, motivation, and instructor's tone or
style. AQ measures concept coverage, cognitive appropriateness, answer
accuracies, and rationale quality. Each lecture's metric score is
computed using the weighted average of all passed rubrics and then
averaged across all 40 lectures generated under each model.

\textbf{Results.} We observed that superior end-to-end performance under a model depends on both the quality of generated lecture contents and the consistency of embodied teaching behavior. As reported in \tabref{tab:frontier-models}, Gemini 3 Pro ranks first overall with an AAR of 80.4\%, supported by strong LCQ (80.2\%), PQ (83.3\%), AQ (81.6\%), and TAQ (76.5\%) scores. GPT-5.1 remains competitive with an AAR of 78.8\%, showing particular strength in Assessment Quality (82.3\%) and Personalization Quality (80.5\%). Claude 4.5 Sonnet follows with the strongest TAQ score (80.4\%), indicating that model choice affects not only content generation but also the quality of embodied teaching actions. 

The rubric breakdown in \figref{fig:rubric-results} further reveals that LCQ limitations across models stem from instruction following and coverage rather than coherence, which remains a relative strength across models. Furthermore, the overall content quality distribution reported in \figref{fig:pq-taq-distribution} shows that structural content rubrics, such as coherence, coverage, cognitive appropriateness, and rationale quality, remain relatively stable across generated artifacts, indicating that LectūraAgents reliably preserves the pedagogical structure of the lecture package. In contrast, the ceiling for personalization remains intrinsically bound to the underlying model, profile sensitivity, and learner's usage of the system into which the framework is built.

\FigureTen
\FigureEleven

\BoldSubsubsection{Lecture Delivery Evaluation}

\noindent Next, we evaluate the
embodied and multimodal teaching capability of the framework. For each
generated lecture, the \emph{ProfessorAgent} is tasked with teaching all
15 slides using lecture artifacts created in the lecture generation
session. This stage evaluates the agent's teaching action quality,
independent of content generation, allowing us to assess multimodal
alignment and embodied instructional delivery capabilities specifically.

\emph{Evaluation Metrics.} Lecture delivery is evaluated using the
Teaching Action Quality (TAQ) metric, which has six rubric dimensions
(detailed in \tabref{tab:evaluation-metrics}). These include temporal and spatial alignment of
teaching actions, accurate handwriting and rough notation actions,
active learning, and overall embodied teaching experience. TAQ assesses
how well each model exploits the frameworks architecture to deliver
accurate, coherent, and pedagogically informed teaching action
sequences. For each slide, script, and teaching action sequence, an
expert educator judges whether each rubric criterion is satisfied, and
the overall average TAQ score is computed using \eqnref{eq:aar}.

\textbf{Results}. TAQ results indicate that LectūraAgents enables generally accurate and coherent teaching action sequences across models. As shown in \figref{fig:rubric-results}, models perform strongly on spatially grounded criteria, particularly spatial accuracy, handwriting actions, rough notation actions, and embodied teaching. This suggests that the framework can reliably convert generated lecture materials into visible instructional actions. \figref{fig:pq-taq-distribution} further shows that teaching-action-related scores are distributed across multiple lecture artifacts, indicating that embodied delivery is maintained across the broader lecture package rather than appearing only in isolated outputs. A key factor behind this stability is the TASA module, which provides the \emph{ProfessorAgent} with a structured view of slide regions and aligns teaching actions with corresponding speech segments. While temporal alignment remains comparatively more variable due to the difficulty of fine-grained action--speech synchronization, \figref{fig:lcq-distribution} shows that TAQ and personalization-related performance remain broadly stable across all learner profiles. This suggests that the embodied teaching mechanism generalizes across academic levels, while timing-sensitive action selection remains an area for improvement.

\BoldSubsubsection{Comparative Evaluation with Related Frameworks} 

\noindent We further assess LectūraAgents against existing
frameworks in this domain. Due to varying capabilities between
baselines, we only compare performances on shared capabilities to
ensure fairness. We identify two closely related open-source frameworks
and one learning system with publicly available lecture data: Instructional Agents
\cite{ref49}, GenMentor
\cite{ref53}, and Google's Learn
Your Way system \cite{ref3}. Our
comparative evaluation assesses each framework or system based on
lecture content quality (LCQ), assessment quality (AQ), and
personalization (PQ) evaluation metrics, using the same evaluation
method described in \secref{sec:experiment-setup}. For InstructionalAgents and GenMentor,
we generated 20 lectures using their publicly released implementations.
For Learn Your Way, we used the publicly available lectures provided on
its website. Additional details about the lecture set and selection
process are provided in \appref{app:code-data}{B}. We then generated the same lectures
with LectūraAgents using identical topics, prompts, and learner
profiles, and evaluated all outputs using the methodology described in
\secref{sec:experiment-setup}.

\TableFive

\emph{Results}. As shown in \tabref{tab:comparison}, LectūraAgents obtains higher scores than the baseline systems across LCQ, PQ, and AQ. The most notable difference is observed in personalization quality, indicating that the framework is better able to adapt generated materials to learner profiles. Its performance in lecture content and assessment quality further suggests that the framework supports not only learner-specific adaptation, but also coherent instructional organization and alignment between lecture materials and assessment tasks.

\TableSix

\BoldSubsubsection{Efficacy Study with Students} 

\noindent The preceding evaluations assessed the pedagogical capabilities of the framework
across multiple topics, models, and personalization settings. However, the
impact of the framework is better examined when these capabilities are
tested on real learners. Therefore, we conducted a small-scale efficacy
study with real students to measure the holistic pedagogical value of
LectūraAgents in terms of learning support and learner experience. To
provide a broader comparison, we included both
Learn Your Way, representing a modern AI-assisted learning system, and
Adobe Acrobat Reader v23.008.20555, representing a widely used
traditional digital study reading software without generative AI
capabilities. The study involved 45 students divided equally across the
three learning systems, with 15 participants per system. Each group comprised five students from each educational level---high school, undergraduate, and master's---with ages ranging between 15 to 25 years. Students were recruited through a short pre-study topic-familiarity screening and provided informed consent prior to participation.

\FigureTwelve

\textbf{Result.} \figref{fig:student-assessment} compares students’ post-learning assessment performance across learning systems. The results show that LectūraAgents achieved the strongest performance across all learner groups, followed by Learn Your Way and Adobe Reader. Although the improvement is modest, its consistency suggests that the framework’s personalized and embodied teaching capabilities supported better short-term comprehension and content recall, rather than merely improving students’ subjective learning experience. Consistent with this pattern, \tabref{tab:student-survey} shows that students using LectūraAgents reported stronger perceived content understanding, assessment readiness, future learning support, and overall learning experience than those using Learn Your Way or Adobe Reader.

\section{Limitations and Future Work}
\label{sec:limitations}

We acknowledge several limitations that may inform future work. First, while LectūraAgents performs well on lecture content generation and embodied delivery, the current teaching action--speech alignment module relies heavily on offline heuristics with a limited set of supported teaching actions. This may constrain the richness of embodied instruction and robustness across diverse slide layouts. Second, the multi-agent orchestration can introduce latency and compute overhead. Finally, the framework can sometimes inherit common LLM failure modes such as factual errors, inconsistent reasoning, and tool or prompt-sensitivity. Future work will (1) expand the teaching action taxonomy and improve action fidelity; (2) transition from heuristic action--speech alignment to learned policies (\eg training policies in a presentation slide environment with preference optimization or reinforcement learning); (3) strengthen grounding to reduce hallucinations; and (4) optimize orchestration for efficiency while preserving pedagogical coherence and controlling compute costs.

\section{Conclusion}
\label{sec:conclusion}

In this paper we introduced LectūraAgents, a hierarchical multi-agent framework for end-to-end adaptive, personalized AI-assisted learning experiences. The framework addresses two major issues in personalized AI-assisted learning: (1) How can AI adaptively personalize instructional contents to best meet the needs of diverse learners? (2) How can such instructional contents be delivered in embodied and pedagogically meaningful ways to ensure better learning outcomes? In order to effectively address these issues, LectūraAgents is first modelled on a professor-student relationship framing, wherein a \emph{ProfessorAgent} leads a collaborative class of specialized subordinate agents through research, planning, evaluation, and embodied delivery of instructional contents that adapt to diverse students. The framework's personalized and embodied capabilities (\eg TASA algorithm) offer students enhanced learning and study experiences. We evaluated LectūraAgents through two main experiments: a pedagogical evaluation under frontier models across high school, undergraduate, and graduate-level topics, and an efficacy study with real students. Experimental results show substantial improvements over baseline frameworks in lecture content quality, personalization, assessment quality, and embodied teaching performance. In addition, these findings are validated by results from our efficacy study with students, which provide preliminary evidence that the framework can improve learning outcomes while enhancing learner experience. In conclusion, we position LectūraAgents offers as a pedagogically grounded framework for personalized AI-assisted learning at scale.

\FloatBarrier

\bibliographystyle{ACM-Reference-Format}
\bibliography{lectura-agents-arxiv}

\clearpage
\onecolumn
\setlength{\parindent}{0pt}
\setlength{\parskip}{0pt plus 0.2pt}
\sloppy
\begingroup
\captionsetup{font={small,stretch=1.08},labelfont=bf,justification=centering,singlelinecheck=false,skip=7pt}
\renewcommand{\arraystretch}{1.12}
\setlength{\tabcolsep}{3pt}
\setlength{\arrayrulewidth}{0.45pt}

\AppendixSection[app:architecture-evaluation]{Appendix A}

\AppendixSubsection[app:detailed-architecture]{A.1 LectūraAgents: Detailed Architecture}

\AppendixSubsubsection{A.1.1 Core Modules and Components}

The framework is organized into four core modules, each serving a
distinct purpose in the lecture generation and lecture delivery stages.
These modules provide the infrastructure for agent coordination, LLM
integration, teaching action alignment, memory management, and content
rendering. The modular design enables easy extension and maintenance of
individual components.

\AppendixTableCaption{Table A1: LectūraAgents' Core Modules and Components}

{\def\LTcaptype{none} 
\begin{longtable}[]{|
  >{\centering\arraybackslash}p{(\linewidth - 6\tabcolsep) * \real{0.1457}}|
  >{\raggedright\arraybackslash}p{(\linewidth - 6\tabcolsep) * \real{0.2960}}|
  >{\raggedright\arraybackslash}p{(\linewidth - 6\tabcolsep) * \real{0.3267}}|
  >{\raggedright\arraybackslash}p{(\linewidth - 6\tabcolsep) * \real{0.2190}}|}
\hline
\begin{minipage}[b]{\linewidth}\centering
\textbf{Module}
\end{minipage} & \begin{minipage}[b]{\linewidth}\centering
\textbf{Location}
\end{minipage} & \begin{minipage}[b]{\linewidth}\centering
\textbf{Function}
\end{minipage} & \begin{minipage}[b]{\linewidth}\centering
\textbf{Key Classes}
\end{minipage} \\
\hline
\endhead
\hline
\endlastfoot
Agents & Lectura/LecturaAgents/module/agents & The Agents module
implements the core agent architecture with base interfaces, role
definitions (Coordinator, Executor, Validator), and state management.
This module provides the hierarchical three-tier agent system with
collaboration mechanisms (sequential and parallel) and orchestration
through SwarmOfRanks. It handles agent lifecycle, task execution,
validation, and inter-agent communication. & Agent (base class),

ProfessorAgent, LecturePlanner,

ResearchAgent,

SlideAgent,

ScriptAgent,

SpeechAgent,

TasaAgent \\
\hline
LLMs & Lectura/LecturaAgents/module/llms & Provides unified abstraction
layer for multiple LLM providers (OpenAI, Google, Anthropic, DeepSeek,
Qwen, Local) enabling seamless model switching. Handles authentication,
API communication, response formatting, function calling, and streaming.
It abstracts provider-specific differences to provide consistent
interface for all agents. & LLMProvider (base class), OpenAIProvider,
GoogleAIProvider, AnthropicProvider, DeepSeekProvider, QwenProvider,
LocalLLMProvider \\
\hline
TASA & Lectura/LecturaAgents/tasa & Implements Teaching Action Salience
Analysis (TASA) for generating and aligning synchronized teaching
actions (rough notation, handwriting) with speech. This module processes
speech scripts to embed action markers, extracts word-level timestamps
from audio using Whisper ASR~\cite{ref66}, and creates temporal alignment between
visual actions and spoken content. & TASA \\
\hline
Adaptive Memory & Lectura/LecturaAgents/memory & Implements a
three-layer adaptive memory system: short-term memory for session
context, long-term memory for persistent learner data, and dynamic
memory for adaptive learning patterns. This module provides a unified
AdaptiveMemory interface that enables agents to access learner context,
preferences, and learning history for personalization. &
ShortTermMemory, LongTermMemory, DynamicMemory, AdaptiveMemory \\
\hline
\end{longtable}
}

\AppendixSubsubsection{A.1.2 Agent Hierarchy and Roles}

Agents are organized by rank and responsibility. Rank 1 agents
(ProfessorAgent) serve as coordinators and validators at the highest
level, Rank 2 agents (LecturePlanner, ResearchAgent) coordinate
execution and validate outputs, while Rank 3 agents (SlideAgent,
ScriptAgent, SpeechAgent, TasaAgent) execute specific tasks. Each agent
has clearly defined responsibilities and access to appropriate tools and
actions for their role.

\clearpage
\AppendixTableCaption{Table A2: Agent Hierarchy and Roles}
{\def\LTcaptype{none} 
\begin{longtable}[]{|
  >{\centering\arraybackslash}p{(\linewidth - 8\tabcolsep) * \real{0.1194}}|
  >{\centering\arraybackslash}p{(\linewidth - 8\tabcolsep) * \real{0.1746}}|
  >{\centering\arraybackslash}p{(\linewidth - 8\tabcolsep) * \real{0.1856}}|
  >{\raggedright\arraybackslash}p{(\linewidth - 8\tabcolsep) * \real{0.3070}}|
  >{\raggedright\arraybackslash}p{(\linewidth - 8\tabcolsep) * \real{0.2018}}|}
\hline
\begin{minipage}[b]{\linewidth}\centering
\textbf{Rank}
\end{minipage} & \begin{minipage}[b]{\linewidth}\centering
\textbf{Agent}
\end{minipage} & \begin{minipage}[b]{\linewidth}\centering
\textbf{Role}
\end{minipage} & \begin{minipage}[b]{\linewidth}\centering
\textbf{Responsibilities}
\end{minipage} & \begin{minipage}[b]{\linewidth}\centering
\textbf{Tools / Actions}
\end{minipage} \\
\hline
\endhead
\hline
\endlastfoot

1 & ProfessorAgent & Coordinator and Tutor &
\begin{minipage}[t]{\linewidth}\raggedright
\begin{itemize}
\item
  Initiates lecture sessions
\item
  Reviews and approves plans
\item
  Validates final artifacts
\item
  Reviews final lecture artifacts
\item
  Delivers embodied lectures
\end{itemize}
\end{minipage} & research()

create\_syllabus()

review\_plan()

instantiate\_groupchat()

create\_lecture\_notes()

create\_study\_guide()

create\_assessments()

create\_personalization()

review\_artifacts()

embodied\_teaching() \\
\hline

2 & Lecture Planner & Validator &
\begin{minipage}[t]{\linewidth}\raggedright
\begin{itemize}
\item
  Creates lecture plans.
\item
  Manages and validates tasks done by subordinate executor agents.
\item
  Assembles generated lecture artifacts and submits to ProfessorAgent
  for final review.
\end{itemize}
\end{minipage} & research()

create\_plan()

validate\_task()

assemble\_artifacts() \\
\hline

\multirow{5}{=}{\centering\arraybackslash 3} &
\multirow{5}{=}{\centering\arraybackslash Executor Agents} &
\multirow{5}{=}{\centering\arraybackslash Executors} &
\textbf{ResearchAgent:} Conducts multi-turn web searches on lecture
topic, writes a detailed research report and submits for review by
LecturePlanner. & web\_search() \\
\cline{4-5}

& & & \textbf{SlideAgent:} Generates personalized slide contents;
Designs and build slides with structured content blocks based on
learner's preferences. & slide\_builder()

research()

file\_parser() \\
\cline{4-5}

& & & \textbf{ScriptAgent:} Creates engaging, personalized narration
scripts that aligns with both slide contents and learner's preferences.
& analyze\_slide()

write\_script() \\
\cline{4-5}

& & & \textbf{SpeechAgent:} Synthesizes and generates speech audio from
scripts based on learner's preferred instructor voice. Uses TTS/ASR tool
to create word-level timestamps . & Whisper~\cite{ref66}, Kokoro TTS~\cite{ref65}

create\_timestamps() \\
\cline{4-5}

& & & \textbf{TasaAgent:} Uses tools in TASA module to segment and
annotate slide contents with heuristic based context for prospective
action-speech sequences. It then processes speech timestamps and slide
contents into synchronized embodied teaching action sequences with
embeded action markers (highlight, underline, handwriting, etc.). & TASA
Module

temporal\_segmentation()

heuristic\_analysis() \\
\hline
\end{longtable}
}

\AppendixSubsubsection{A.1.3 Agent States and Lifecycle}

Agents transition through a well-defined state machine during task execution. The lifecycle begins with the IDLE state, progresses through acknowledgment and execution phases, and concludes with completion, failure, or revision states. This state management ensures proper task tracking, error handling, and enables agents to revise their work based on feedback from higher-ranking agents.

\clearpage
\AppendixTableCaption{Table A3: LectūraAgents' States and Lifecycle}

{\def\LTcaptype{none} 
\begin{longtable}[]{|
  >{\centering\arraybackslash}p{(\linewidth - 4\tabcolsep) * \real{0.1816}}|
  >{\raggedright\arraybackslash}p{(\linewidth - 4\tabcolsep) * \real{0.2766}}|
  >{\raggedright\arraybackslash}p{(\linewidth - 4\tabcolsep) * \real{0.3093}}|}
\hline
\begin{minipage}[b]{\linewidth}\centering
\textbf{State}
\end{minipage} & \begin{minipage}[b]{\linewidth}\centering
\textbf{Description}
\end{minipage} & \begin{minipage}[b]{\linewidth}\centering
\textbf{Transition}
\end{minipage} \\
\hline
\endhead
\hline
\endlastfoot
IDLE & Agent is waiting for a task. & ACKNOWLEDGED \\
\hline
ACKNOWLEDGED & Agent has received and acknowledged given. & EXECUTING \\
\hline
EXECUTING & Agent is actively working on assigned task. & COMPLETED,
FAILED or REVISAL \\
\hline
COMPLETED & Agent has completed task successfully. & IDLE (for next
task) \\
\hline
FAILED & Task execution was unsuccessful. & REVISAL \\
\hline
REVISAL & Agent is revising work based on feedback from self-reflection
or review. & EXECUTING \\
\hline
\end{longtable}
}

\AppendixSubsubsection{A.1.4 Multi-agent Collaboration}

Agents within the same rank can collaborate using two primary
mechanisms: sequential collaboration for dependent tasks and parallel
collaboration for independent tasks. The SwarmOfRanks mechanism enables
hierarchical coordination across multiple ranks, allowing complex
workflows where agents at different levels coordinate their activities.
These collaboration patterns are essential for orchestrating the
multi-stage lecture generation process.

\AppendixTableCaption{Table A4: Collaboration Mechanisms}

{\def\LTcaptype{none} 
\begin{longtable}[]{|
  >{\centering\arraybackslash}p{(\linewidth - 4\tabcolsep) * \real{0.1886}}|
  >{\raggedright\arraybackslash}p{(\linewidth - 4\tabcolsep) * \real{0.2623}}|
  >{\centering\arraybackslash}p{(\linewidth - 4\tabcolsep) * \real{0.2534}}|}
\hline
\begin{minipage}[b]{\linewidth}\centering
\textbf{Type}
\end{minipage} & \begin{minipage}[b]{\linewidth}\centering
\textbf{Description}
\end{minipage} & \begin{minipage}[b]{\linewidth}\centering
\textbf{Use Case}
\end{minipage} \\
\hline
\endhead
\hline
\endlastfoot
Sequential Colab & Agents complete tasks one after another, sharing
responses. & When tasks depend on previous outputs. \\
\hline
Parallel Colab & Agents complete tasks simultaneously, while sharing
responses. & When tasks are independent. \\
\hline
Swarm of Ranks & Hierarchical coordination across ranks & Multi-rank
workflows \\
\hline
\end{longtable}
}

\AppendixSubsubsection{A.1.5 Tools and Capabilities}

The framework provides a comprehensive set of tools that agents use to
accomplish their tasks. These tools range from web search and file
parsing to text-to-speech synthesis and code execution. Each tool is
designed to be modular and reusable, with clear interfaces that agents
can invoke during their execution. The tools abstract away complex
operations like API interactions, file processing, and multimedia
generation.

\AppendixTableCaption{Table A5: Tools and Capabilities}

{\def\LTcaptype{none} 
\begin{longtable}[]{|
  >{\centering\arraybackslash}p{(\linewidth - 6\tabcolsep) * \real{0.1370}}|
  >{\raggedright\arraybackslash}p{(\linewidth - 6\tabcolsep) * \real{0.2730}}|
  >{\raggedright\arraybackslash}p{(\linewidth - 6\tabcolsep) * \real{0.3100}}|
  >{\centering\arraybackslash}p{(\linewidth - 6\tabcolsep) * \real{0.2726}}|}
\hline
\begin{minipage}[b]{\linewidth}\centering
\textbf{Tool}
\end{minipage} & \begin{minipage}[b]{\linewidth}\centering
\textbf{Purpose}
\end{minipage} & \begin{minipage}[b]{\linewidth}\centering
\textbf{Usage}
\end{minipage} & \begin{minipage}[b]{\linewidth}\centering
\textbf{Dependencies}
\end{minipage} \\
\hline
\endhead
\hline
\endlastfoot
Web Search & Multi-turn web research using SerpAPI. & Used by
\emph{ResearchAgent, ProfessorAgent, LecturePlanner} and
\emph{SlideAgent} & SerpAPI \\
\hline
Slide World & Dynamic slide environment with canvas for teaching
sessions. & Used by \emph{ProfessorAgent} for embodied lecture delivery.
& HTML/CSS/JS/Python \\
\hline
Slide Builder & Custom slide design tool. & Used by \emph{SlideAgent}
for building and rendering slides. & HTML/CSS/JS/Python \\
\hline
File Parser & Parses PDF, TXT, MD files. & Used by \emph{ProfessorAgent}
and \emph{SlideAgent} to extract content from additional materials. &
PyPDF2, python-docx \\
\hline
Command line & For command execution to create lecture artifacts. & Used
by all agents to read/write/edit/save/delete files. & Bash/Zsh \\
\hline
TASA Segmentor / Aligner & Segments, annotates and aligns slide contents
with speech timestamps for synchronized & Used by \emph{ProfessorAgent}
and \emph{TasaAgent} & TASA Module \\
\hline
Research & A unified research tool that makes use of web search plus an
LLM to perform deep research on topics. & Used by \emph{ResearchAgent,
ProfessorAgent, LecturePlanner} and \emph{SlideAgent} & SerpAPI +
Underlying LLM \\
\hline
Whisper~\cite{ref66} & Extracts word-level timestamps from audio. & Used by
\emph{TasaAgent} for action alignment. & Whisper ASR model \\
\hline
Kokoro TTS~\cite{ref65} & Generate speeches from scripts with desired instructor
voice. & Used by \emph{SpeechAgent} for speech synthesis. & Kokoro
TTS \\
\hline
\end{longtable}
}

\AppendixSubsubsection{A.1.6 Adaptive Memory}

LectūraAgents utilizes a three-layer memory architecture to support
adaptive and personalized learning experiences. Short-term memory
captures recent interactions within a session, long-term memory stores
persistent learner-specific data across sessions, and dynamic memory
adapts to learning patterns and preferences. The adaptive memory module
provides a unified interface that combines all three memory types,
enabling agents to access relevant context efficiently.

\AppendixTableCaption{Table A6: Memory Types and Functionalities}

{\def\LTcaptype{none} 
\begin{longtable}[]{|
  >{\centering\arraybackslash}p{(\linewidth - 6\tabcolsep) * \real{0.1888}}|
  >{\raggedright\arraybackslash}p{(\linewidth - 6\tabcolsep) * \real{0.2626}}|
  >{\centering\arraybackslash}p{(\linewidth - 6\tabcolsep) * \real{0.2536}}|
  >{\centering\arraybackslash}p{(\linewidth - 6\tabcolsep) * \real{0.2936}}|}
\hline
\begin{minipage}[b]{\linewidth}\centering
\textbf{Memory Type}
\end{minipage} & \begin{minipage}[b]{\linewidth}\centering
\textbf{Function}
\end{minipage} & \begin{minipage}[b]{\linewidth}\centering
\textbf{Storage}
\end{minipage} & \begin{minipage}[b]{\linewidth}\centering
\textbf{Update Frequency}
\end{minipage} \\
\hline
\endhead
\hline
\endlastfoot
Short-term Memory & Handles recent interactions and context. & In-memory
(session-based) & Per interaction \\
\hline
Long-term Memory & Manages persistent learner-specific data. &
File-based (JSON) & Per session \\
\hline
Dynamic Memory & Adaptive learning patterns and preferences. & In-memory
+ file-based & Continuously updated \\
\hline
\end{longtable}
}

\AppendixSubsubsection{A.1.7 LLMs}

We ensure the framework supports multiple frontier models from leading
LLM providers through a unified API, allowing seamless switching between
different models based on task requirements, cost considerations, and
performance needs. Each provider implementation handles authentication,
API communication, and response formatting, while the unified interface
ensures that agents can work with any supported model without code
changes. This design enables flexibility in choosing the most
appropriate model for each task.

\AppendixTableCaption{Table A7: Supported LLM Providers and Models}

{\def\LTcaptype{none} 
\begin{longtable}[]{|
  >{\centering\arraybackslash}p{(\linewidth - 2\tabcolsep) * \real{0.2308}}|
  >{\centering\arraybackslash}p{(\linewidth - 2\tabcolsep) * \real{0.2805}}|}
\hline
\begin{minipage}[b]{\linewidth}\centering
\textbf{LLM Provider}
\end{minipage} & \begin{minipage}[b]{\linewidth}\centering
\textbf{Supported Models}
\end{minipage} \\
\hline
\endhead
\hline
\endlastfoot
OpenAI & GPT-5.1, GPT-4o, o3-pro \\
\hline
Google AI & Gemini 3 Pro, Gemini 2.5 Pro, Gemini Flash 2.5 Lite \\
\hline
Anthropic & Claude 4.5 Sonnet, Claude 4.1 Sonnet \\
\hline
DeepSeek & DeepSeek V3.2, DeepSeek-R1 \\
\hline
\end{longtable}
}

\AppendixSubsubsection{A.1.8 Slide Content Block Types}

To ensure accurate alignment and rubust slide contents, we ensure each
slide can support multiple content block types that enable rich,
structured presentation of information. Each block type is designed for
specific pedagogical purposes, from definitions and equations for core
concepts to examples, steps, and questions for engagement. The framework
automatically renders these blocks with appropriate styling and
formatting, ensuring consistent visual presentation across all slides.

\AppendixTableCaption{Table A8: Various Types of Slide Content Blocks}

{\def\LTcaptype{none} 
\begin{longtable}[]{|
  >{\centering\arraybackslash}p{(\linewidth - 6\tabcolsep) * \real{0.1888}}|
  >{\raggedright\arraybackslash}p{(\linewidth - 6\tabcolsep) * \real{0.2626}}|
  >{\centering\arraybackslash}p{(\linewidth - 6\tabcolsep) * \real{0.2536}}|
  >{\centering\arraybackslash}p{(\linewidth - 6\tabcolsep) * \real{0.2936}}|}
\hline
\begin{minipage}[b]{\linewidth}\centering
\textbf{Block Type}
\end{minipage} & \begin{minipage}[b]{\linewidth}\centering
\textbf{Description}
\end{minipage} & \begin{minipage}[b]{\linewidth}\centering
\textbf{Rendering}
\end{minipage} & \begin{minipage}[b]{\linewidth}\centering
\textbf{Usage}
\end{minipage} \\
\hline
\endhead
\hline
\endlastfoot
Bullets & Brief, concise key points about concepts and topics. & HTML
list elements (\textless ul\textgreater\textless/ul\textgreater,
\textless ol\textgreater\textless/ol\textgreater, etc.) & Holds main
contents for topic \\
\hline
Definition & Key term definitions. & HTML styled definition div & Core
concepts \\
\hline
Example & Concrete examples & HTML highlighted example div & Examples \\
\hline
Equation & Mathematical equations & LaTeX rendering in a div & Formulas,
proofs \\
\hline
Question & Interactive questions & HTML Question box div & Engagement \\
\hline
Link & External references & hyperlink / link element & Resources \\
\hline
Table & Structured data & HTML table element & Comparisons, data \\
\hline
Video & YouTube video embeds & HTML iframe element & Educational short
videos \\
\hline
Image & Illustrative and educative images with captions & HTML image
element & Illustration \\
\hline
Steps & Step-by-step procedures & HTML numbered list & Algorithms,
processes, etc. \\
\hline
\end{longtable}
}

\AppendixSubsection[app:evaluation-methodology]{A.2 More on Evaluation Methodology}

\AppendixSubsubsection{A.2.1 Overview}

Our evaluation adopts a rubric-based methodology for both pedagogical
and comparative assessment, with generated learning and teaching
artifacts scored and validated by expert educators. The evaluation
examines two core capabilities of the framework: its ability to generate
high-quality personalized lecture content for diverse learner profiles,
and its ability to utilize these generated materials during embodied
teaching. Specifically, we evaluate LectūraAgents using four main
metrics: Lecture Content Quality (LCQ), Personalization
Quality (PQ), Assessment Quality (AQ), and Teaching
Action Quality (TAQ). These metrics are applied across three evaluation
settings: (1) Pedagogical Evaluation under Frontier Models,
which assesses personalized lecture generation and embodied lecture
delivery across different frontier models; (2) Comparative
Evaluation with Related Frameworks, which compares LectūraAgents with
existing educational agent or personalized learning frameworks,
including InstructionalAgents, LearnYourWay, and GenMentor; and
(3) Efficacy Study with Students, which examines the
framework's practical learning support and learner experience using real
student participants.

\AppendixSubsubsection[app:pedagogical-eval]{A.2.2 LectūraAgents' Pedagogical Evaluation Under Frontier Models}

During this evaluation, we generated 40 lectures per model across seven
models, resulting in 280 lectures in total. For each model, the lecture
set included 10 lectures per academic level, with 20 learner profiles in
total (five profiles per level). The topics covered science,
engineering, history, art, and business. Details on these lectures can
be found in the released dataset, available at HuggingFace\footnote{
HuggingFace dataset:~\url{https://huggingface.co/datasets/Jaward/lectura-agents-data}}. The
generated lecture artifacts were assessed across four evaluation
metrics: Lecture Content Quality (LCQ), Personalization Quality (PQ),
Assessment Quality (AQ), and Teaching Action Quality (TAQ). The
evaluation followed a two-stage procedure. In Stage 1, an LLM analyst
provided structured rubric-based analysis for each lecture, identifying
evidence relevant to the instructional criteria under each metric, as
detailed in \apptabref{tab:appendix-stages}{A9} and \apptabref{tab:appendix-metrics}{A10}. In Stage 2, expert educators reviewed the
LLM-generated analysis, validated the evidence, assigned the final
rubric scores, and made corrections where necessary. The verified scores
were then aggregated to compute metric-level scores, overall averages,
visualizations, and comparative insights into model performance across
academic levels and evaluation dimensions.

\AppendixTableCaption[tab:appendix-stages]{Table A9: Stages in Pedagogical Evaluations}

{\def\LTcaptype{none} 
\begin{longtable}[]{|
  >{\centering\arraybackslash}p{(\linewidth - 6\tabcolsep) * \real{0.1888}}|
  >{\raggedright\arraybackslash}p{(\linewidth - 6\tabcolsep) * \real{0.2626}}|
  >{\raggedright\arraybackslash}p{(\linewidth - 6\tabcolsep) * \real{0.2536}}|
  >{\centering\arraybackslash}p{(\linewidth - 6\tabcolsep) * \real{0.2936}}|}
\hline
\begin{minipage}[b]{\linewidth}\centering
\textbf{Stage}
\end{minipage} & \begin{minipage}[b]{\linewidth}\centering
\textbf{Task}
\end{minipage} & \begin{minipage}[b]{\linewidth}\centering
\textbf{Command}
\end{minipage} & \begin{minipage}[b]{\linewidth}\centering
\textbf{Output}
\end{minipage} \\
\hline
\endhead
\hline
\endlastfoot
Stage 1 & An LLM (GPT 5.2) gives detail analysis of generated lecture contents per academic level based on rubrics or criteria in the evaluation metrics. & python3 evaluate.py \textbackslash{}

-\/-model model\_name \textbackslash{}

-\/-lecture lecture\_name \textbackslash{}

-\/-level level\_name \textbackslash{}

-\/-llm analysis\_model & (JSON)

Detailed analysis for each generated lecture at each academic level under a model. \\
\hline
Stage 2 & An expert educator validates, scores and aggregate results for respective rubrics. & python3 evaluate.py \textbackslash{}

-\/-aggregate \textbackslash{}

-\/-lecture lecture\_name \textbackslash{}

-\/-level level\_name & (JSON, Charts)

Comprehensive scores and results. \\
\hline
\end{longtable}
}

\AppendixTableCaption[tab:appendix-metrics]{Table A10: Details on Evaluation Metrics, Rubrics, Descriptions and Their Input Files}

{\def\LTcaptype{none} 
\begin{longtable}[]{|
  >{\centering\arraybackslash}p{(\linewidth - 6\tabcolsep) * \real{0.1604}}|
  >{\centering\arraybackslash}p{(\linewidth - 6\tabcolsep) * \real{0.2480}}|
  >{\centering\arraybackslash}p{(\linewidth - 6\tabcolsep) * \real{0.3091}}|
  >{\centering\arraybackslash}p{(\linewidth - 6\tabcolsep) * \real{0.2826}}|}
\hline
\multicolumn{4}{|c|}{\textbf{Lecture Generation Evaluation}}\\
\hline
\textbf{Evaluation Metric} & \textbf{Rubrics / Criteria} & \textbf{Description} &
\textbf{Input Files} \\
\hline
\endfirsthead

\hline
\textbf{Evaluation Metric} & \textbf{Rubrics / Criteria} & \textbf{Description} &
\textbf{Input Files} \\
\hline
\endhead

\hline
\endfoot

\hline
\endlastfoot

Lecture Content Quality (LCQ) & \emph{Accuracy} & Verifies factual correctness across all generated materials. & All generated files \\
\cline{2-4}
& \emph{Clarity} & Assesses clarity of explanation across teaching materials. & lecture\_plan.json, learner\_profile.txt, syllabus.json, scripts.json, slides\_content.json, slides/*.html, lecture\_notes\_/*.md, quiz.json, and exam.json \\
\cline{2-4}
& \emph{Coherence} & Evaluates logical flow across all materials. & All generated files \\
\cline{2-4}
& \emph{Cognitive Load} & Assesses lecture contents alignment with learner's background or level. & learner\_profile.txt, syllabus.json, scripts.json, slides\_content.json, slides/*.html, lecture\_notes\_/*.md, quiz.json, and exam.json \\
\cline{2-4}
& \emph{Syllabus Coverage} & Verifies topic coverage. & syllabus.json, scripts.json, slides\_content.json, slides/*.html, lecture\_notes\_/*.md, quiz.json, exam.json, and study\_guide.md \\
\cline{2-4}
& \emph{Instruction-following} & Checks framework's adherence to instructions, tasks or prompts. & All generated files \\
\hline

Personalization Quality (PQ) & \emph{Adaptive Emphasis} & Assesses the framework's ability to adapt instructions to the learner's learning preferences or profile through. & learner\_profile.txt, scripts.json, slides\_content.json, slides/*.html, lecture\_notes\_/*.md, quiz.json, exam.json, and study\_guide.md \\
\cline{2-4}
& \emph{Preference Alignment} & Checks content alignment with learning
preferences. & teaching\_actions.json, scripts.json,
slides\_content.json, slides/*.html, lecture\_notes\_/*.md, quiz.json, exam.json, and study\_guide.md \\
\cline{2-4}
& \emph{Engagement} & Evaluates framework's capability to consistently
engage the learner. & teaching\_actions.json, scripts.json,
slides\_content.json, slides/*.html, lecture\_notes\_/*.md, quiz.json, exam.json, and study\_guide.md \\
\cline{2-4}
& \emph{Motivation} & Evaluate motivational elements across learning materials. & teaching\_actions.json, scripts.json, slides\_content.json, slides/*.html, lecture\_notes\_/*.md, quiz.json, exam.json, and study\_guide.md \\
\cline{2-4}
& \emph{Tone/Style} & Evaluate language appropriateness & scripts.json, slides\_content.json, lecture\_notes\_/*.md, study\_guide.md, and learner\_profile.txt \\
\hline

Assessment Quality (AQ) & \emph{Concept Coverage} & Verifies whether assessments covered all
topics in the syllabus. & quiz.json, exam.json, syllabus.json,
slides\_content.json \\
\cline{2-4}
& \emph{Cognitive Appropriateness} & Evaluates assessment difficulty and
its alignment with the learner's profile. & learner\_profile.txt,
quiz.json, exam.json, syllabus.json, slides\_content.json \\
\cline{2-4}
& \emph{Answer Validity} & Checks accuracy of solutions to assessments. &
quiz.json, exam.json, syllabus.json, slides\_content.json \\
\cline{2-4}
& \emph{Rationale} & Evaluates the quality of explanation in solutions. &
quiz\_solutions.json, exam\_solutions.json \\
\hline

\multicolumn{4}{|c|}{\textbf{Lecture Delivery Evaluation}}\\
\hline
\textbf{Evaluation Metric} & \textbf{Rubrics / Criteria} & \textbf{Description} &
\textbf{Input Files} \\
\hline

Teaching Action Quality (TAQ) & \emph{Temporal Alignment} & Validates action-speech alignments. & action\_speech\_alignment.json, scripts.json, speech\_timestamps.json \\
\cline{2-4}
& \emph{Accurate Handwriting Action} & Checks accuracy of handwriting actions, \ie whether words or phrases are written clearly and correctly at the right time frame. & slides/*.html (after applied actions), action\_speech\_alignment.json \\
\cline{2-4}
& \emph{Accurate Rough Notation} & Checks accuracy of rough notation actions, \ie whether notations like highlight, underline, and circle actions are applied correctly in the right region and at the right time frame. & slides/*.html (after applied actions,
action\_speech\_alignment.json \\
\cline{2-4}
& \emph{Spatial Accuracy} & Verifies annotation precision. & slides/*.html (after applied actions),
action\_speech\_alignment.json \\
\cline{2-4}
& \emph{Active Learning} & Assesses the effect of teaching actions on the learner's engagement or focus during teaching. & slides/*.html, quiz.json, exam.json, action\_speech\_alignment.json \\
\cline{2-4}
& \emph{Embodied Teaching} & Evaluates overall embodied teaching experience. & tasa\_analysis.json, teaching\_actions.json, slides/*.html (after applied actions), action\_speech\_alignment.json, scripts.json, speech\_timestamps.json \\
\hline
\end{longtable}
}

\AppendixSubsubsection[app:rating]{A.2.3 Rating}

Each rubric or criteria is evaluated as a boolean (satisfied or not), and these boolean scores are weighted and averaged to produce Average Achieved Ratings (AARs) at the metric and overall levels. Thus, for the j-th lecture, the overall performance score for under a given model, is computed as the weighted average of all passed rubric criteria \({AAR}_{w}^{j}\), given by:

\[
\begin{gathered}
{AAR}_{w}^{j} =
\frac{\sum_{i = 1}^{N_j} w_i^j \cdot \mathbf{1}_{r_i^j}}
{\sum_{i = 1}^{N_j} w_i^j \cdot \mathbf{1}_{w_i^j > 0}}
\end{gathered}
\]

where \(N_{j}\) is the number of rubric criteria for the j-th lecture, \(w_{i}^{j}\)\ensuremath{\in} \{-5, -3, -1, 0, +1, +3, +5\}, is the weight assigned to the i-th criterion and \(r_{i}^{j}\) \ensuremath{\in} \{0,1\}indicates whether criterion i is satisfied. When a criterion is satisfied \(r_{i}^{j} = 1\), it contributes a positive reward of +5, +3 or +1, corresponding to a highly desirable, a desirable and important, or a nice-to-have behaviour, respectively. When a criterion is not satisfied \(r_{i}^{j} = 0\), it is explicitly treated as a failure state and contributes a non-positive score, spanning 0, -1, -3, -5: 0 denotes the lowest-severity failure (no credit), -1 a minor failure, -3 a moderate failure, and -5 a critical failure (highly undesirable behaviour).

\AppendixSubsubsection[app:expert-recruitment]{A.2.4 Expert Recruitment and Evaluation Procedure}

Five expert educators were recruited through purposive sampling based on their experience in teaching, curriculum development, and educational assessment. The panel consisted of secondary-school teachers and university instructors from STEM, social science, and humanities disciplines, each with at least five years of teaching experience. Prior to the evaluation, the experts participated in an online workshop, during which the evaluation dimensions, criteria, and weighting scheme were reviewed and refined to ensure pedagogical relevance and consistency across educational levels and subject domains. During the evaluation, experts were assigned respective lecture samples according to their areas of expertise; they reviewed the generated lecture artifacts and assigned final scores based on the agreed-upon rubrics.

\AppendixSubsubsection[app:comparative-eval]{A.2.5 Comparative Evaluation of LectūraAgents with Related Frameworks}

Comparative analysis was done against two multi-agent frameworks (Instructional Agents and GenMentor) and one system (Google's Learn Your Way). For the frameworks, we generated 20 lectures (5 for each level spanning 10 profiles) using their released code and then generated the same lectures with LectūraAgents and compared performances. \apptabref{tab:appendix-comparative}{A11} summarizes generated lecture topics and profiles per framework or system. For Google's Learn Your Way system, given that no source code was released we instead utilized their already generated sample lectures openly available on their website. We then generated these lectures with LectūraAgents and compared performances as well. Our comparative evaluation assesses each framework or system based on lecture content quality (LCQ), assessment quality (AQ) and personalization (PQ) evaluation metrics using the same evaluation method  described in \appref{app:rating}{A.2.3} and \appref{app:expert-recruitment}{A.2.4}.

\AppendixTableCaption[tab:appendix-comparative]{Table A11: Generated Lectures for Comparative Analysis}

{\def\LTcaptype{none} 
\begin{longtable}[]{|
  >{\raggedright\arraybackslash}p{(\linewidth - 2\tabcolsep) * \real{0.2246}}|
  >{\raggedright\arraybackslash}p{(\linewidth - 2\tabcolsep) * \real{0.7739}}|}
\hline
\begin{minipage}[b]{\linewidth}\centering
\textbf{Framework / System}
\end{minipage} & \begin{minipage}[b]{\linewidth}\centering
\textbf{Lecture and Learner Profile Details}
\end{minipage} \\
\hline
\endhead
\hline
\endlastfoot

Instructional Agents and\newline GenMentor & Lecture Title: \emph{Newton's Laws of
Motion}

Learner Profile: \emph{8th-grade high schooler interested in STEM,
enjoys basketball, and prefers visual, hands-on learning through
diagrams, examples, and practical activities.} \\
\cline{2-2}
& \begin{minipage}[t]{\linewidth}\raggedright
Lecture Title: \emph{Photosynthesis and Cellular Respiration}\\
Learner Profile: \emph{9th-grade high schooler interested in creative
writing and music, enjoys sketching, and learns biology best through
story-like explanations, visuals, and everyday analogies.}\strut
\end{minipage} \\
\cline{2-2}
& \begin{minipage}[t]{\linewidth}\raggedright
Lecture Title: \emph{Quadratic Equations and Functions}\\
Learner Profile: \emph{10th-grade high schooler preparing for advanced
mathematics, enjoys chess, and prefers worked examples, graph-based
explanations, and short practice problems.}\strut
\end{minipage} \\
\cline{2-2}
& \begin{minipage}[t]{\linewidth}\raggedright
Lecture Title: \emph{The Solar System and Planetary Motion}\\
Learner Profile: \emph{11th-grade high schooler interested in astronomy
and planetary systems, enjoys tennis, and prefers simulations, diagrams,
and applied problem solving.}\strut
\end{minipage} \\
\cline{2-2}
& \begin{minipage}[t]{\linewidth}\raggedright
Lecture Title: \emph{World War II: Causes and Consequences}\\
Learner Profile: \emph{12th-grade high schooler interested in modern
history and global conflict, enjoys soccer, and prefers timeline-based
explanations with cause-and-effect reasoning.}\strut
\end{minipage} \\
\cline{2-2}
& \begin{minipage}[t]{\linewidth}\raggedright
Lecture Title: \emph{Intro to Large Language Models}\\
Learner Profile: \emph{Undergraduate computer science student interested
in artificial intelligence and language technologies, enjoys basketball,
and prefers intuitive explanations followed by coding examples.}\strut
\end{minipage} \\
\cline{2-2}
& \begin{minipage}[t]{\linewidth}\raggedright
Lecture Title: \emph{Machine Learning: Supervised vs Unsupervised}\\
Learner Profile: \emph{Undergraduate data science student interested in
machine learning methods and data patterns, enjoys hiking, and prefers
visual comparisons using real datasets.}\strut
\end{minipage} \\
\cline{2-2}
& \begin{minipage}[t]{\linewidth}\raggedright
Lecture Title: \emph{Molecular Biology: Gene Expression}\\
Learner Profile: \emph{Undergraduate biology student interested in
genetics and molecular regulation, enjoys swimming, and prefers process
diagrams with concept checks.}\strut
\end{minipage} \\
\cline{2-2}
& \begin{minipage}[t]{\linewidth}\raggedright
Lecture Title: \emph{Operating Systems: Process Scheduling}\\
Learner Profile: \emph{Undergraduate learner interested in environmental
science and sustainability, enjoys photography, and learns systems
concepts best through visual workflows, resource-allocation analogies,
and practical examples.}\strut
\end{minipage} \\
\cline{2-2}
& \begin{minipage}[t]{\linewidth}\raggedright
Lecture Title: \emph{Thermodynamics: Entropy and Free Energy}\\
Learner Profile: \emph{Undergraduate chemistry student interested in
thermodynamics and energy transformations, enjoys cooking, and prefers
equation walkthroughs connected to everyday examples.}\strut
\end{minipage} \\
\cline{2-2}
& \begin{minipage}[t]{\linewidth}\raggedright
Lecture Title: \emph{Advanced Machine Learning: Deep Neural Networks}\\
Learner Profile: \emph{Master's-level engineering
student interested in deep learning and neural architectures, enjoys
tennis, and prefers model diagrams with optimization intuition.}\strut
\end{minipage} \\
\cline{2-2}
& \begin{minipage}[t]{\linewidth}\raggedright
Lecture Title: \emph{Advanced Operating Systems}\\
Learner Profile: \emph{Master's-level systems student
interested in distributed computing and resource management, enjoys
cycling, and prefers architecture diagrams with performance
trade-offs.}\strut
\end{minipage} \\
\cline{2-2}
& \begin{minipage}[t]{\linewidth}\raggedright
Lecture Title: \emph{Computational Biology: Sequence Analysis}\\
Learner Profile: \emph{Master's-level computational
biology student interested in genomics and sequence alignment, enjoys
photography, and prefers algorithmic workflows with biological
examples.}\strut
\end{minipage} \\
\cline{2-2}
& \begin{minipage}[t]{\linewidth}\raggedright
Lecture Title: \emph{Cryptography and Network Security}\\
Learner Profile: \emph{Master's-level learner interested
in ancient history and ethics, enjoys debate, and learns cryptography
best through historical examples, trust scenarios, and clear protocol
diagrams.}\strut
\end{minipage} \\
\cline{2-2}
& \begin{minipage}[t]{\linewidth}\raggedright
Lecture Title: \emph{Distributed Systems Architecture}\\
Learner Profile: \emph{Master's-level computer science
student interested in scalable systems and fault tolerance, enjoys
tennis, and prefers system-design scenarios with failure cases..}\strut
\end{minipage} \\
\cline{2-2}
& \begin{minipage}[t]{\linewidth}\raggedright
Lecture Title: \emph{Advanced Quantum Field Theory}\\
Learner Profile: \emph{PhD researcher interested in quantum fields and
particle interactions, enjoys baseball, and prefers formal derivations
supported by physical intuition.}\strut
\end{minipage} \\
\cline{2-2}
& \begin{minipage}[t]{\linewidth}\raggedright
Lecture Title: \emph{Non-Equilibrium Statistical Mechanics}\\
Learner Profile: \emph{PhD researcher interested in statistical physics
and complex systems, enjoys tennis, and prefers rigorous mathematical
development with simulation examples.}\strut
\end{minipage} \\
\cline{2-2}
& Lecture Title: \emph{Synthetic Biology: Circuit Design}

Learner Profile: \emph{PhD researcher interested in synthetic biology
and programmable cellular circuits, enjoys running, and prefers circuit
schematics with lab-oriented examples.} \\
\cline{2-2}
& \begin{minipage}[t]{\linewidth}\raggedright
Lecture Title: \emph{Topological Data Analysis in ML}\\
Learner Profile: \emph{PhD researcher interested in topology and machine
learning geometry, enjoys rock climbing, and prefers visual abstractions
grounded in data examples.}\strut
\end{minipage} \\
\hline

Learn Your Way & Lecture Title: \emph{Atoms and
Molecules}

Learner Profile: Middle schooler who likes reading. \\
\cline{2-2}
& \begin{minipage}[t]{\linewidth}\raggedright
Lecture Title: \emph{Carbon}\\
Learner Profile: \emph{Undergrad who likes painting.}\strut
\end{minipage} \\
\cline{2-2}
& \begin{minipage}[t]{\linewidth}\raggedright
Lecture Title: \emph{Microeconomics and Macroeconomics}\\
Learner Profile: \emph{Undergrad who likes food.}\strut
\end{minipage} \\
\cline{2-2}
& \begin{minipage}[t]{\linewidth}\raggedright
Lecture Title: \emph{Logical Statements}\\
Learner Profile: \emph{Undergrad who likes writing.}\strut
\end{minipage} \\
\cline{2-2}
& \begin{minipage}[t]{\linewidth}\raggedright
Lecture Title: \emph{The Ancient Roman Economy}\\
Learner Profile: \emph{Undergraduate who likes plants..}\strut
\end{minipage} \\
\cline{2-2}
& \begin{minipage}[t]{\linewidth}\raggedright
Lecture Title: \emph{The \textquotesingle Long-Haired\textquotesingle{}
Comets}\\
Learner Profile: \emph{Undergraduate who likes movies..}\strut
\end{minipage} \\
\cline{2-2}
& \begin{minipage}[t]{\linewidth}\raggedright
Lecture Title: \emph{Early Human Evolution and Migration}\\
Learner Profile: \emph{Undergrad who like tennis..}\strut
\end{minipage} \\
\cline{2-2}
& \begin{minipage}[t]{\linewidth}\raggedright
Lecture Title: \emph{Intro to Data Structures and Algorithms}\\
Learner Profile: \emph{High schooler who likes basketball.}\strut
\end{minipage} \\
\cline{2-2}
& Lecture Title: \emph{Critical Reading and Evidence-Based Response}

Learner Profile: \emph{Middle schooler who likes soccer.} \\
\cline{2-2}
& \begin{minipage}[t]{\linewidth}\raggedright
Lecture Title: \emph{Disruptions in the Immune System}\\
Learner Profile: \emph{Middle schooler who likes food}\strut
\end{minipage} \\
\cline{2-2}
& \begin{minipage}[t]{\linewidth}\raggedright
Lecture Title: \emph{Earth and Sky}\\
Learner Profile: \emph{Middle schooler who likes photography}\strut
\end{minipage} \\
\cline{2-2}
& \begin{minipage}[t]{\linewidth}\raggedright
Lecture Title: \emph{Theories of Slef-development}\\
Learner Profile: \emph{Undergrad who likes cooking.}\strut
\end{minipage} \\
\cline{2-2}
& \begin{minipage}[t]{\linewidth}\raggedright
Lecture Title: \emph{What is Learning\\
}Learner Profile: \emph{Undergrad who likes music.}\strut
\end{minipage} \\
\cline{2-2}
& \begin{minipage}[t]{\linewidth}\raggedright
Lecture Title: \emph{''Reading'' to Understand and respond}\\
Learner Profile: \emph{Middle schooler who likes music.}\strut
\end{minipage} \\
\cline{2-2}
& \begin{minipage}[t]{\linewidth}\raggedright
Lecture Title: \emph{Micronomics and Macronomics}\\
Learner Profile\emph{: Undergrad who likes cooking.}\strut
\end{minipage} \\
\cline{2-2}
& \begin{minipage}[t]{\linewidth}\raggedright
Lecture Title: \emph{An Overview of Economic Systems\\
}Learner Profile: \emph{High schooler who likes movies.}\strut
\end{minipage} \\
\cline{2-2}
& \begin{minipage}[t]{\linewidth}\raggedright
Lecture Title: \emph{Early Human Evolution and Migration\\
}Learner Profile: \emph{Undergrad who likes tennis}\strut
\end{minipage} \\
\hline
\end{longtable}
}

\clearpage
\AppendixSection[app:code-data]{Appendix B}

\AppendixSubsection[app:code-data-section]{B.1 Code and Data}

The data supporting this study is currently available on our huggingface
repository at:
https://huggingface.co/datasets/Jaward/lectura-agents-data. The code can
be made available upon reasonable request from the corresponding author.
Please follow the installation instructions below or in the readme file
to get started.

\AppendixSubsubsection[app:installation]{B.1.1 Installation and Usage}

\par\vskip -0.1\baselineskip
\AppendixNumberedItem{1}{Add all required api keys inside the .env file in the parent
  directory. You will need to provide two main api keys (1) for the LLM
  you want to use (OpenAI, Anthropic, Gemini and Deepseek); (2) A
  SerpApi key for research, while this is optional, it highly
  recommended to add one, as it helps reduce hallucination. Get key
  here: https://serpapi.com/manage-api-key}
\AppendixNumberedItem{2}{Cd into the parent directory and install all required packages using this command:}
\AppendixCommandBlock{pip3 install -r requirements.txt}
\AppendixNumberedItem{3}{If you wish to use the frontend for lecture generation, start the app
  with this command:}
\AppendixCommandBlock{python3 main.py}
\begin{figure}[H]
\centering
\includegraphics[width=0.8\textwidth]{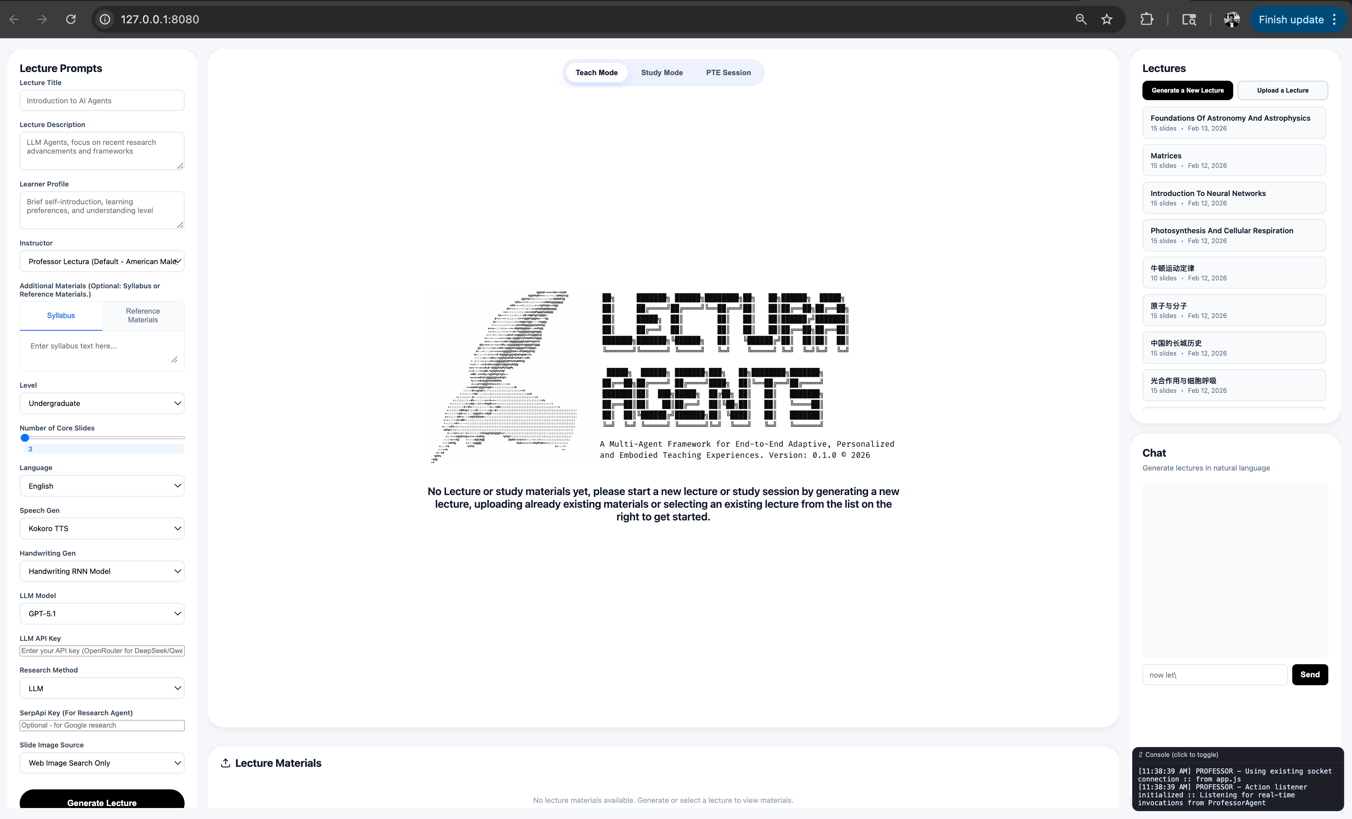}
\caption{Frontend view (with no generated lecture)}
\Description{This figure is described in the caption.}
\label{fig:appendix-frontend}
\end{figure}
This
will open the teaching environment in your browser at:
\url{http://127.0.0.1:8080/}. The page should look like \figref{fig:appendix-frontend}:

There will be a few already generated lectures in the right Lectures
pane for you to quickly try or you can also generate new lectures
through either the chat pane or in the left prompt pane. Generated
lecture materials will appear below the slide as they are generated.

\begin{figure}[H]
\centering
\includegraphics[width=0.8\textwidth]{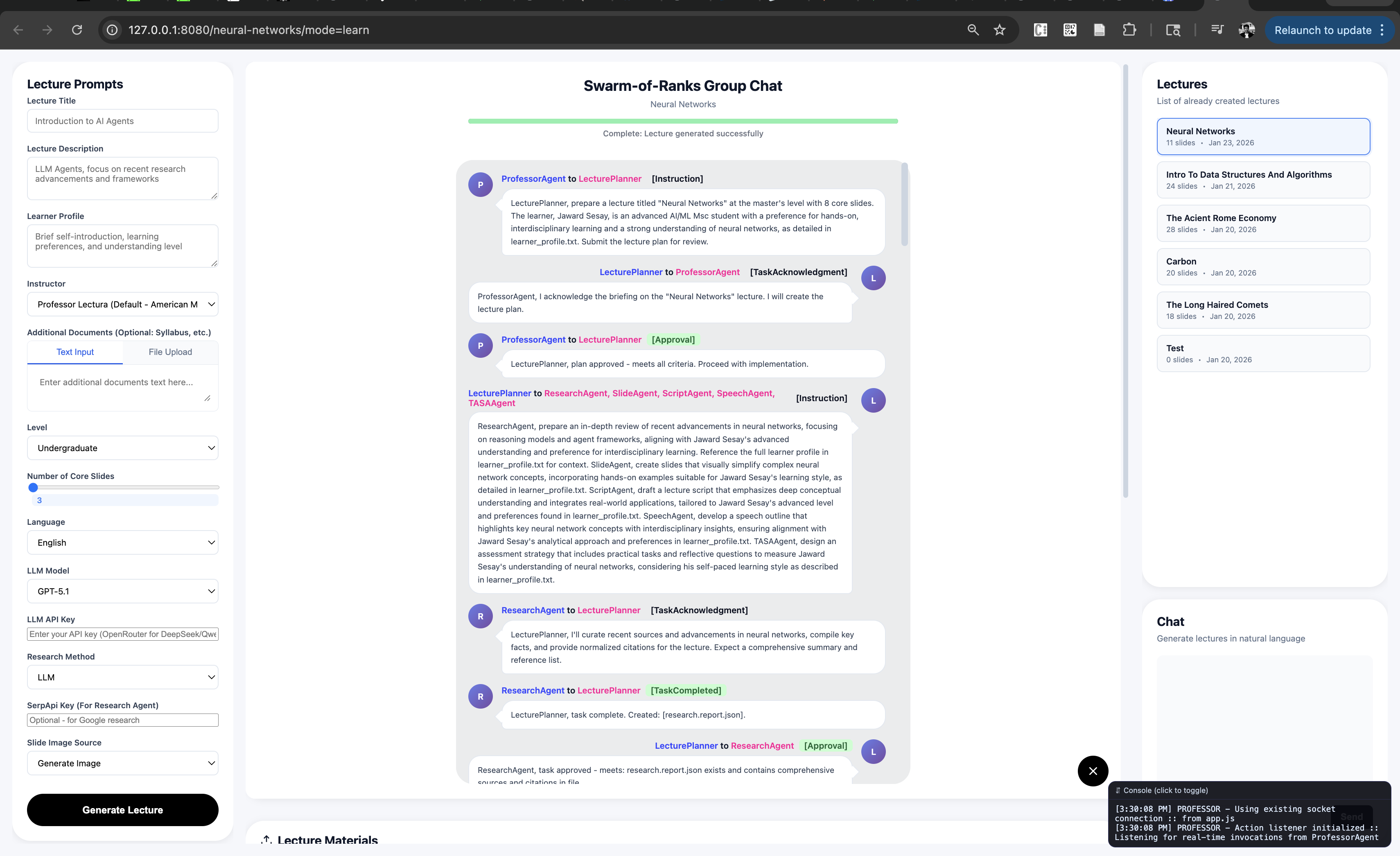}
\caption{Swarm-of-Ranks group chat view (during lecture generation)}
\Description{This figure is described in the caption.}
\label{fig:appendix-swarm}
\end{figure}
During
lecture generation you can follow the whole process unfolds in real-time
in the group chat session, as shown in \figref{fig:appendix-swarm}.

\begin{figure}[H]
\centering
\includegraphics[width=0.8\textwidth]{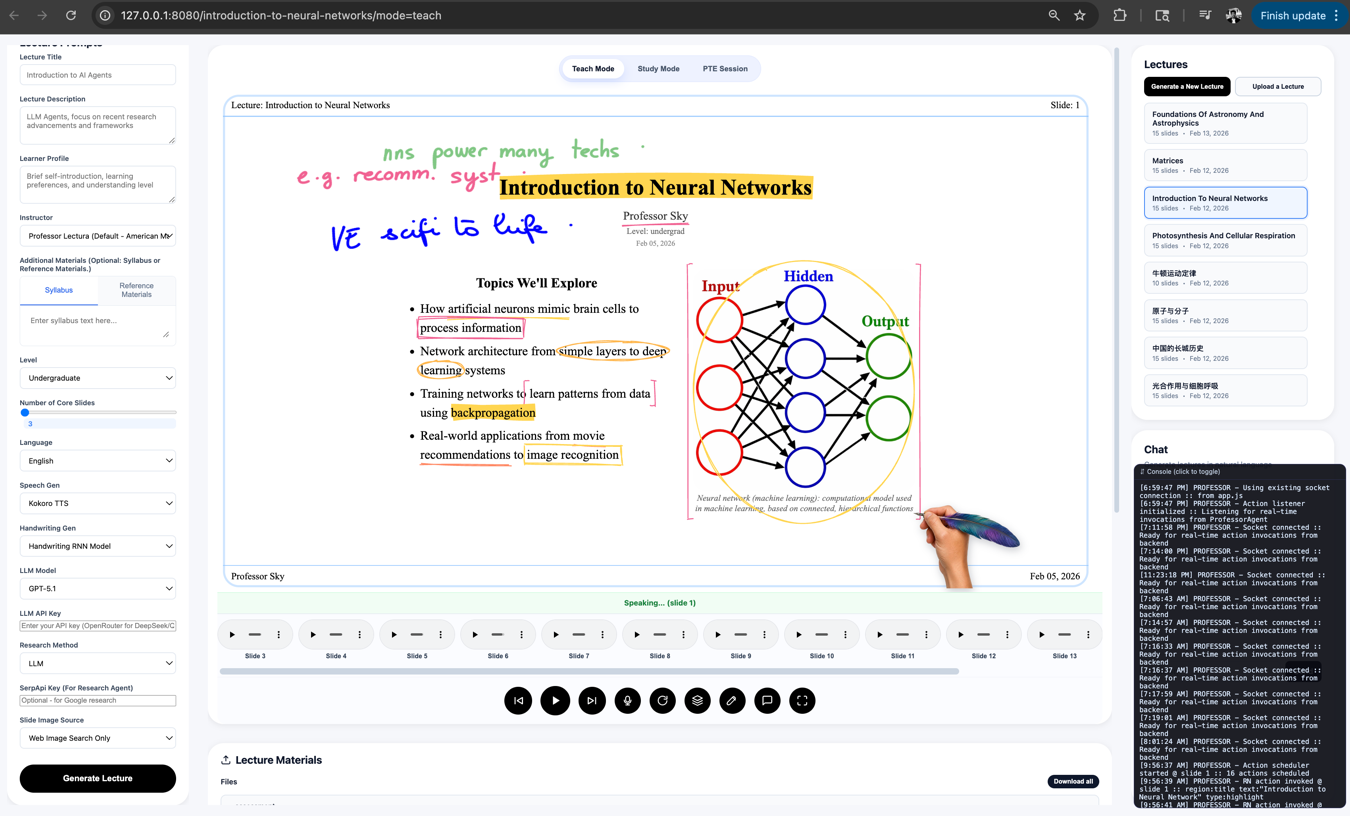}
\caption{Teaching and learning environment (with teaching actions on lecture or study contents)}
\Description{This figure is described in the caption.}
\label{fig:appendix-teaching-env}
\end{figure}

After lecture generation is complete, the view will automatically update with the slide deck (as shown in \figref{fig:appendix-teaching-env}). Below the deck are controls (Next, Play, Previous, Restart, Temporal Segmentation, and Chat).

\AppendixNumberedItem{4}{If you wish to use the terminal for lecture generation, run this command:}
\AppendixCommandBlock{python3 lecture\_prep.py \textbackslash{}\\
-\/-lecture\_title "Your Lecture Title Goes Here"
\textbackslash{}\\
-\/-lecture\_desc "Describe the kind of lecture you want here"
\textbackslash{}\\
-\/-learner\_profile "Add details about yourself, your learning
preferences, and your current understanding level here"
\textbackslash{}\\
-\/-slides \textless enter desired number of core slides
here\textgreater{} \textbackslash{}\\
-\/-level \textless enter academic level: highschool, undergrad,
masters, or phd\textgreater{} \textbackslash{}\\
-\/-instructor\_voice \textless choose desired instructor voice:
professor\_lectura, professor\_sky, professor\_isabella, etc..
\textgreater{} \textbackslash{}\\
-\/-llm \textless select desired model here: gpt-5.1, gpt-4o, o3-pro,
gemini-3-pro, gemini-flash-2.5-lite, claude-4.5,
claude-4.1\textgreater{} \textbackslash{}\\
-\/-research \textless enter research method: llm or
google\textgreater{} \textbackslash{}\\
-\/-language \textless enter output language: english, chinese, french,
or spanish\textgreater{} \textbackslash{}\\
-\/-speech\_gen \textless choose speech backend: kokoro-tts,
gemini-2.5-tts, or gpt-4o-mini-tts\textgreater{} \textbackslash{}\\
-\/-handwriting\_gen \textless choose handwriting mode:
handwriting\_rnn\_model or preset\_font\_handwriting\textgreater{}
\textbackslash{}\\
-\/-slide\_image \textless choose slide image mode: generate\_only,
generate\_web\_search, web\_search\_only, material\_generate\_alt,
material\_web\_alt, or material\_only\textgreater{} \textbackslash{}\\
-\/-syllabus "Optional syllabus or curriculum text here"
\textbackslash{}\\
-\/-additional\_materials "Optional reference text or path(s) to .pdf,
.txt, or .md files, separated by commas" \textbackslash{}\\
-\/-data\_root \textless optional custom output directory\textgreater{}}
\AppendixNumberedItem{5}{Example prompt:}
\AppendixCommandBlock{python3 lecture\_prep.py \textbackslash{}\\
-\/-lecture\_title "Intro to Data Structures and Algorithms" \textbackslash{}\\
-\/-lecture\_desc "A Computer Science lecture for a highschooler who likes basketball. Ensure covering these topics and more: 1. Introduction to Data Types and Abstraction 2. Introduction to Algorithms 3. Algorithm Vs Program. Understanding Data Structures 4. Abstract Data Types: (List, Set, Map, Priority Queue, Graph)" \textbackslash{}\\
-\/-learner\_profile "Name: Taylor. Focus: Advanced Computer Science. Interests: Specialized algorithms, system design. Hobby: basketball. Learning style: Deep dive into technical details." \textbackslash{}\\
-\/-slides 24 \textbackslash{}\\
-\/-level highschool \textbackslash{}\\
-\/-instructor\_voice professor\_sky \textbackslash{}\\
-\/-llm gpt-5.2 \textbackslash{}\\
-\/-research google}
\AppendixNumberedItem{6}{To view the generated lecture in the teaching environment run this command:}
\AppendixCommandBlock{python3 lecture\_delivery.py -\/-lecture \textless lecture folder name\textgreater{}}

The folder could be, for example, \textit{intro-to-data-structures-and-algorithms}.
\endgroup
\end{document}